\documentclass[10pt]{article} 
\usepackage[preprint]{tmlr}

\usepackage{amsmath,amsfonts}
\usepackage{algorithmic}
\usepackage{algorithm}
\usepackage{array}
\usepackage[caption=false,font=normalsize,labelfont=sf,textfont=sf]{subfig}
\usepackage{textcomp}
\usepackage{stfloats}
\usepackage{url}
\usepackage{verbatim}
\usepackage{graphicx}
\usepackage{rotating}
\usepackage[utf8]{inputenc}
\usepackage[T1]{fontenc}
\usepackage{hyperref}
\usepackage{booktabs}
\usepackage{nicefrac}
\usepackage{tabularx}
\usepackage{multirow}
\usepackage{microtype}
\usepackage{xcolor}
\usepackage{soul}
\usepackage{adjustbox}
\hypersetup{
    colorlinks=true,
    linkcolor=blue,
    citecolor=red,
    urlcolor=cyan,
}
\usepackage{tikz}
\usetikzlibrary{positioning,arrows}
\usepackage{forest}
\definecolor{lightblue}{rgb}{.90,.95,1}
\sethlcolor{lightblue}

\usepackage[capitalize,nameinlink,noabbrev]{cleveref}
\usepackage{enumitem}

\title{Emergent Abilities in Large Language Models: A Survey}

\author{\name Leonardo Berti \email leonardo.berti@evolvingdata.net \\
      EvolvingData
      \AND
      \name Flavio Giorgi \email giorgi@di.uniroma1.it \\
      Sapienza University of Rome
      \AND
      \name Gjergji Kasneci \email gjergji.kasneci@tum.de \\
      Technical University of Munich
}

\begin{document}

\maketitle

\begin{abstract}
Large Language Models (LLMs) are leading a new technological revolution as one of the most promising research streams toward artificial general intelligence. The scaling of these models, accomplished by increasing the number of parameters and the magnitude of the training datasets, has been linked to various so-called \emph{emergent abilities} that were previously unobserved. These emergent abilities, ranging from advanced reasoning and in-context learning to coding and problem-solving, have sparked an intense scientific debate: \emph{Are they truly emergent, or do they simply depend on external factors, such as training dynamics, the type of problems, or the chosen metric? What underlying mechanism causes them?} Despite their transformative potential, emergent abilities remain poorly understood, leading to misconceptions about their definition, nature, predictability, and implications.
In this work, we shed light on emergent abilities by conducting a comprehensive review of the phenomenon, addressing both its scientific underpinnings and real-world consequences. We first critically analyze existing definitions, exposing inconsistencies in conceptualizing emergent abilities. We then explore the conditions under which these abilities appear, evaluating the role of scaling laws, task complexity, pre-training loss, quantization, and prompting strategies. Our review extends beyond traditional LLMs and includes Large Reasoning Models (LRMs), which leverage reinforcement learning and inference-time search to amplify reasoning and self-reflection. However, emergence is not inherently positive. As AI systems gain autonomous reasoning capabilities, they also develop harmful behaviors, including deception, manipulation, and reward hacking. We highlight growing concerns about safety and governance, emphasizing the need for better evaluation frameworks and regulatory oversight.
\end{abstract}


\section{Introduction}\label{sec:intro}

The study of emergent properties in complex systems has been a long-standing interdisciplinary pursuit, spanning fields such as physics, biology, and mathematics. While the term emergent was coined by G. H. Lewes \citep{lewes1877problems} in 1877, the concept of emergence gained widespread recognition through  Anderson's seminal work, ``\textit{More Is Different}'' \cite{anderson1972more}.  Anderson postulated that, as systems increase in complexity, novel surprising properties may manifest, even with a comprehensive quantitative understanding of their microscopic constituents. 
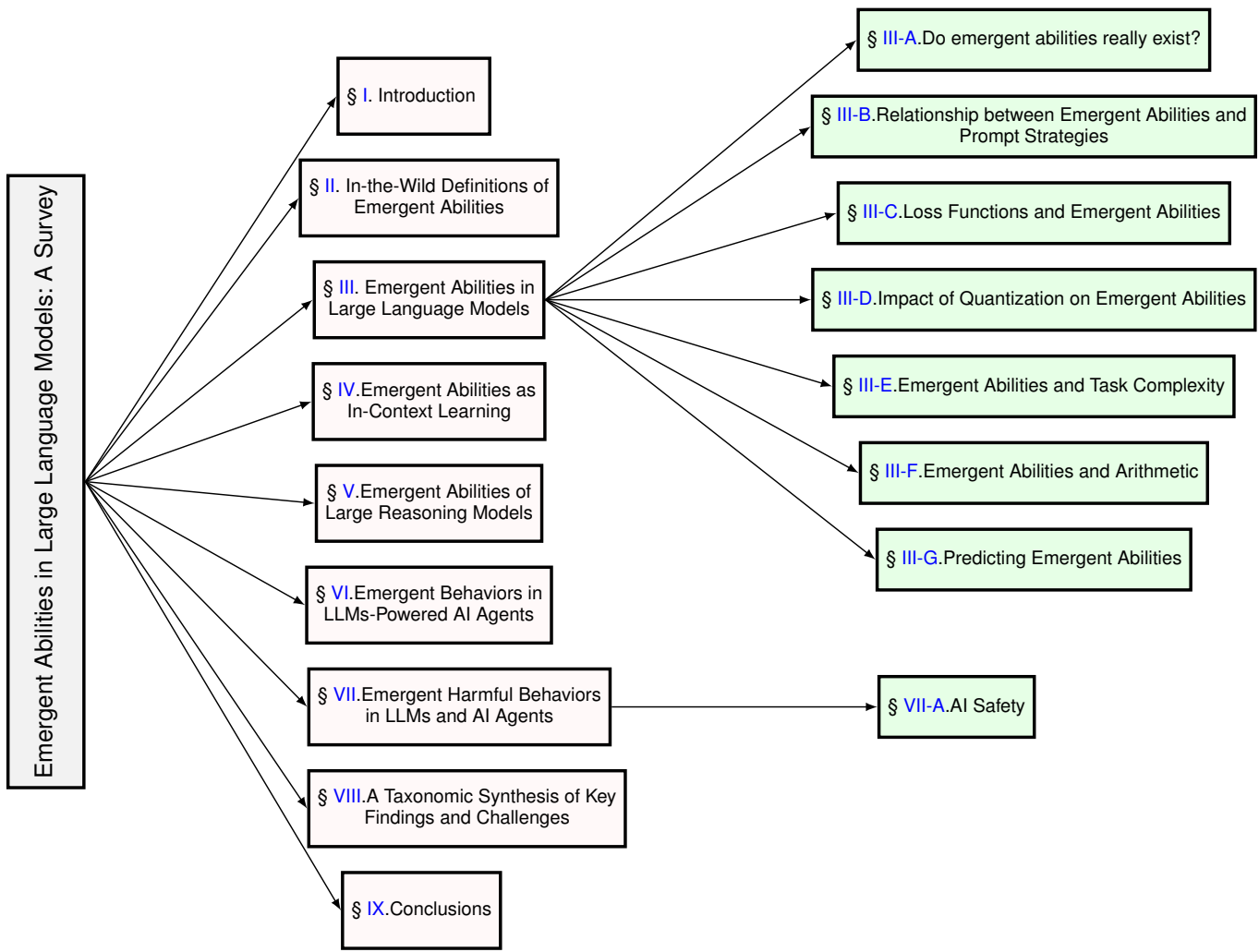
\begin{figure*}
    \centering
    \resizebox{\textwidth}{!}{%
        \begin{tikzpicture}
            [
                node distance=0.26cm and 3.5cm,
                title/.style={
                    rectangle,
                    draw,
                    fill=gray!10,
                    very thick,
                    minimum width=1cm,
                    minimum height=1cm,
                    font=\sffamily\small,
                    rotate=90
                },
                sections/.style={
                    align=center,
                    rectangle,
                    draw,
                    fill=pink!10,
                    very thick,
                    minimum width=1.5cm,
                    minimum height=0.94cm,
                    font=\sffamily\scriptsize
                },
                subsections/.style={
                    align=center,
                    rectangle,
                    draw,
                    fill=green!10,
                    very thick,
                    minimum width=1cm,
                    minimum height=0.8cm,
                    font=\sffamily\scriptsize
                },
                line/.style={
                    -latex,
                    very thick,
                    line width=0.5pt
                }
            ]


            \node[title, anchor=center] (E)
            {Emergent Abilities in Large Language Models: A Survey};


            \node[
                sections,
                right=of E,
                yshift=-2.7cm
            ] (D)
            {§~\ref{sec:in_context}. Emergent Abilities as\\
             In-Context Learning};

            \node[
                sections,
                above=of D,
                yshift=0.03cm
            ] (C)
            {§~\ref{sec:emergent_abilities_llms}. Emergent Abilities in\\
             Large Language Models};

            \node[
                sections,
                above=of C,
                yshift=0.03cm
            ] (B)
            {§~\ref{sec:definitions}. In-the-Wild Definitions of\\
             Emergent Abilities};

            \node[
                sections,
                above=of B
            ] (A)
            {§~\ref{sec:intro}. Introduction};

            \node[
                sections,
                below=of D,
                yshift=0.1cm
            ] (F)
            {§~\ref{sec:lrms}. Emergent Abilities of\\
             Large Reasoning Models};

            \node[
                sections,
                below=of F
            ] (G)
            {§~\ref{sec:llm_agents}. Emergent Behaviors in\\
             LLMs-Powered AI Agents};

            \node[
                sections,
                below=of G
            ] (H)
            {§~\ref{sec:harmful_behaviors}. Emergent Harmful Behaviors\\
             in LLMs and AI Agents};

            \node[
                sections,
                below=of H
            ] (conc)
            {§~\ref{sec:conclusion}. Conclusions};


            \node[
                subsections,
                right=of C,
                yshift=-0.8cm
            ] (quant)
            {§~\ref{subsec:quant}. Impact of Quantization on Emergent Abilities};

            \node[
                subsections,
                above=of quant
            ] (loss)
            {§~\ref{subsec:loss}. Loss Functions and Emergent Abilities};

            \node[
                subsections,
                above=of loss
            ] (prompt)
            {§~\ref{subsec:prompt}. Relationship between Emergent Abilities and\\
             Prompt Strategies};

            \node[
                subsections,
                above=of prompt
            ] (exist)
            {§~\ref{sec:mirage}. Do emergent abilities really exist?};

            \node[
                subsections,
                below=of quant
            ] (complex)
            {§~\ref{subsec:complex}. Emergent Abilities and Task Complexity};

            \node[
                subsections,
                below=of complex
            ] (arith)
            {§~\ref{subsec:arith}. Emergent Internal Representations and Mechanisms};

            \node[
                subsections,
                below=of arith
            ] (predict)
            {§~\ref{subsec:predict}. Predicting Emergent Abilities};


            \node[
                subsections,
                right=of H,
                yshift=-0.5cm
            ] (decept)
            {§~\ref{subsec:safety}. AI Safety};


            \draw[line] (E.south) -- (A.west);
            \draw[line] (E.south) -- (B.west);
            \draw[line] (E.south) -- (C.west);
            \draw[line] (E.south) -- (D.west);
            \draw[line] (E.south) -- (F.west);
            \draw[line] (E.south) -- (G.west);
            \draw[line] (E.south) -- (H.west);
            \draw[line] (E.south) -- (conc.west);


            \draw[line] (C.east) -- (exist.west);
            \draw[line] (C.east) -- (prompt.west);
            \draw[line] (C.east) -- (loss.west);
            \draw[line] (C.east) -- (quant.west);
            \draw[line] (C.east) -- (complex.west);
            \draw[line] (C.east) -- (arith.west);
            \draw[line] (C.east) -- (predict.west);


            \draw[line] (H.east) -- (decept.west);

        \end{tikzpicture}%
    }

    \caption{Overview of Emergent Abilities in Large Language Models}
    \label{fig:emergent_abilities}
\end{figure*}

This paradigm shift challenges the constructionist approach, which consists of reconstructing and understanding complex systems solely through the extrapolation of individual particle properties. Anderson prescribes the development of alternative laws that can capture the holistic nature of emergent phenomena in complex systems. Ten years later, Hopfield~\cite{hopfield1982neural} marked the inception of the concept of emergent abilities in neural networks. Drawing parallels from physical systems comprised of numerous simple elements, he observed that collective phenomena, such as stable magnetic orientations or vortex patterns in fluid dynamics, arise from the interactions of these basic elements.  This observation prompted Hopfield to investigate whether the computational capabilities of neural networks could be understood as an emergent property resulting from the interactions of many simple neuronal units. Anderson's and Hopfield's insights laid the foundation for understanding how complex behavior can emerge from simple interactions, a principle that continues to influence modern artificial neural networks. This idea has become particularly relevant in deep learning with the advent of large language models (LLMs) \citep{brown2020language, touvron2023llama, zhao2023survey}. These models have fundamentally revolutionized the field of natural language processing, achieving state-of-the-art performance through novel techniques such as in-context learning and chain-of-thought prompting. By leveraging a few examples within the input prompt, LLMs demonstrate a remarkable ability to generalize to new tasks without explicit fine-tuning. Not only do these models exhibit improved performance, but they also demonstrate unexpected behaviors, giving rise to emergent abilities that were not anticipated or present in smaller models.  
The correlation between the scale of language models, as measured by training compute and model parameters, and their efficacy in various downstream natural language processing (NLP) tasks has been well established in the literature \citep{devlin2018bert, brown2020language}. The impact of scale on model performance can frequently be predicted through empirically derived scaling laws \citep{kaplan2020scaling, hoffmann2022training}, whose mechanistic origin has been recently linked to representation superposition in LLM embeddings \citep{liu2025superposition}. However, these relationships are not universally applicable. Intriguingly, certain downstream tasks exhibit a discontinuous relationship between model scale and performance, unpredictably defying the general trend of continuous improvement. This phenomenon underscores the complexity inherent in the scaling dynamics of language models and highlights the need for new approaches to understanding and predicting their behavior across various applications.

Understanding emergent abilities in LLMs is fundamental to ensuring system reliability and safety, particularly in predicting the emergence of harmful capabilities \citep{perez2022ignorepreviouspromptattack, bai2022training, hagendorff2024deception, williams2024targeted}, such as manipulation and the dissemination of misinformation.

This work comprehensively reviews the study of emergent abilities for LLMs. Despite growing academic and public interest in this phenomenon, there remains a lack of consensus on the precise definition of emergent abilities, leading to significant conceptual ambiguity and confusion. Ironically, the need for clarification on emergent properties is not novel. Johnson et al.~\citep{johnson2006emergent} previously addressed this topic, although they focused on their manifestation in the engineering of complex systems.
To address this definitional uncertainty, we first conduct a systematic analysis of current definitions in the literature (\cref{sec:definitions}). 
Then we comprehensively review the literature on emergent abilities (\cref{sec:emergent_abilities_llms}) as defined by Wei et al. \citep{wei2022emergent}.
Subsequently, we examine in-context learning (\cref{sec:in_context}) and continue by exploring the emergent abilities of Large Reasoning Models (LRMs), a new class of AI systems that extend traditional LLMs by incorporating reinforcement learning post-training and inference-time search (\cref{sec:lrms}). 
Beyond beneficial emergent abilities, we also examine the rise of LLM-powered AI agents and their implications (\cref{sec:llm_agents}). 
We analyze the emergence of harmful behaviors in LLMs and LLM-powered agents (\cref{sec:harmful_behaviors}), as advanced AI systems have demonstrated deceptive tendencies and reinforcement learning-driven manipulation that could lead to unintended consequences. 

\paragraph{Scope and limitations of this survey.}
Our focus is emergence in text-based LLMs and the reasoning models and agents built on them:
its competing definitions (\cref{sec:definitions}), the conditions under which it appears---scale,
pre-training loss, task complexity, quantization, and prompting (\cref{sec:emergent_abilities_llms})---its relationship to in-context learning (\cref{sec:in_context}), its mechanistic underpinnings, and the
harmful behaviors that emerge alongside beneficial ones (\cref{sec:harmful_behaviors}). Several
boundaries are deliberate. We do not systematically treat \emph{multimodal} or vision-language
emergence, invoking multimodal results only where they illuminate mechanisms shared with text-only
models; emergence in non-linguistic domains (e.g., protein, code-only, or reinforcement-learning
agents outside the LLM setting) is likewise out of scope. We draw on mechanistic
interpretability~\citep{bereska2024mechanistic} selectively, in service of the mechanism-over-metric
stance, rather than surveying it as a field, and we treat scaling
laws~\citep{kaplan2020scaling, hoffmann2022training} and AI governance (\cref{subsec:safety}) at the
level needed to situate emergence rather than as standalone reviews. We make no attempt to re-enumerate
every reported emergent task, deferring to the 137 abilities catalogued by Wei et
al.~\citep{wei2022emergent}; instead we prioritize works that clarify \emph{why} and \emph{when}
abilities emerge and \emph{whether} an internal mechanism accompanies them. Papers were identified by
keyword search over the emergent-abilities literature combined with citation-graph tracing from the
foundational works, favoring studies that add mechanistic, statistical, or predictive insight over
those that only report a new benchmark jump. Finally, this is a fast-moving field: the specific
model-and-benchmark figures we cite (e.g., frontier scores on GSM8K, GPQA, AIME, and ARC-AGI) are
snapshots that will date quickly, even as the structural findings we emphasize are likely to persist.

\section{In-the-Wild Definitions of Emergent Abilities}\label{sec:definitions}
We begin by examining various definitions in the literature, proceeding from general conceptualizations to LLM-specific definitions, thereby providing a hierarchical and historical framework for understanding this phenomenon.

The earliest definition can be found in the work by George Henry Lewes in 1877~\cite{lewes1877problems}:

\begin{quote}
``Each stage of evolution presents itself as the consequence of a preceding stage, \hl{at once an emergence and a continuance}; so that \hl{no transposition of stages is possible}; each has its appointed place'' \\\\\hspace*{\fill}-- George Henry Lewes
\end{quote}

Lewes emphasizes that evolution proceeds in a fixed, sequential order. Each stage is both a continuation of previous developments and an emergence of new properties arising from those earlier stages.

Later definitions focus more on the complexity of systems. 

\begin{quote}
``The behavior of large and complex aggregates of elementary particles, it turns out, is not to be understood in terms of a simple extrapolation of the properties of a few particles. Instead, \hl{at each level of complexity entirely new properties appear}, and the \hl{understanding} of the \hl{new behaviors} requires research...'' \\\\\hspace*{\fill}-- Philip W. Anderson 
\end{quote}

Anderson's definition \citep{anderson1972more} is universal and concerns complex systems in general. Anderson's definition also refers to the fields of science, dividing the world into different strata. Interestingly, Anderson proposes a layered view of complexity\footnote{We can consider levels of complexity as the orders of magnitude of the parameters and training compute in the case of LLMs.} and emergence. At the bottom are fundamental physical laws. Climbing the hierarchy, we can observe chemistry, biology, psychology, and social sciences. Anderson states that emergent abilities are those properties that (1) emerge at each level of complexity and (2) cannot be understood simply by analyzing the single components' behavior. For instance, the phenomenon of life, as studied in biology, is an emergent property of chemistry and physics. Recently, we have witnessed an enormous surge in different LLMs from several stakeholders, e.g., LLaMa (Meta), Claude (Anthropic), GPT (OpenAI), and Gemini (Google).  Moreover, the same stakeholders push towards bigger models regarding parameter number and training compute. Interestingly, Anderson's definition of emergent abilities holds true in the LLM era, especially with models containing more and more parameters. 

\begin{quote}
``\hl{Computational properties} of use to biological organisms or the construction of computers can \hl{emerge as collective properties} of systems having a \hl{large number of simple} equivalent components (or \hl{neurons})''. \\\\\hspace*{\fill}-- John J Hopfield
\end{quote}

Hopfield~\citep{hopfield1982neural} tackles emergent properties in simple-structured neural networks with neurons having elementary properties. Here, he notices that collective computational properties spontaneously arise where the so-called ``memories'' are stable entities or \textit{Gestalts} that can be correctly accessed from any network subpart.  He posits that the architecture of animal brains is made from the interconnection of simple local circuits with well-defined functions (à la activations in artificial neural networks). This makes the collective behavior of large quantities of simple processing elements sprout to the spontaneous emergence of new computational capabilities.
This can be considered the most general definition.
Fast-forward to the LLM era, notice how Hopfield's observations encompass all the computational tasks LLMs can perform. Although emergent abilities might surface from multiple simple local circuits of neurons, concluding whether an individual neuron is conscious is short-sighted. Interestingly, Li et al.~\citep{li2024memory} suggest a potential correspondence between Tulving's synergistic ecphory model of retrieval and the emergent abilities observed in LLMs. Nevertheless, arguing whether full-fledged LLMs are sentient or self-aware is an open challenge with many pro et contra points of view \citep{DBLP:journals/corr/abs-2303-07103}.

\begin{quote}
``An ability is emergent if it is \hl{not present in smaller} models \hl{but} is present \hl{in larger models}. Emergent abilities \hl{would not have been directly predicted} by extrapolating a scaling law from small-scale models. When visualized via a scaling curve, emergent abilities show a clear pattern- performance is near-random until a
certain critical threshold of scale is reached, after which \hl{performance increases to substantially above random}.'' \\\\\hspace*{\fill}-- Jason Wei et al.
\end{quote}
This definition is the first large language model-specific definition and is the most used in the academic literature \citep{schaeffer2024emergent, ganguli2022predictability, lu2023emergent}. With respect to the two previous definitions, it adds two important concepts: i.e., the \textbf{unpredictability} and the \textbf{magnitude} of the performance increase. Additionally, this definition entails that LLMs, after reaching a specific scale, perform better than random for them to be considered emergent.
\begin{center}
\begin{quote}
``Emergent abilities are \hl{in-context learning}.''  -- Tacit consensus in the media
\end{quote}
\end{center}
The term emergent in the LLM context is also used simply to indicate all the capabilities that develop \textbf{implicitly} during next-token prediction-based pre-training. These abilities are tested with few-shot prompting without gradient updates to the model. This process is referred to as in-context learning and is the capability to generalize from a few examples to new \textbf{tasks} and \textbf{concepts} on which they have not been directly trained. This observation can also be generalized to zero-shot prompting.
\begin{quote}
``\hl{Emergence should} \hl{require at minimum the identification of relevant coarse-grained variables that form effective mechanisms}---reduced ``internal degrees of freedom''---for this behavior, mechanisms that can explain or predict the behavior of the system at this higher level, screening off details of lower level mechanisms such as weights and activations.'' \\\\\hspace*{\fill}-- D.C. Krakauer, J.W. Krakauer \& M. Mitchell
\end{quote}
The most recent reframing comes from Krakauer, Krakauer and Mitchell~\cite{krakauer2025complex}, who situate LLM emergence within the complexity-science tradition initiated by Anderson and Hopfield. They argue that the prevailing LLM-specific definition---centred on abrupt benchmark jumps with scale---is incomplete because it omits the constitutive feature of emergence in physics and biology: the system must reorganise so that a \textbf{new, lower-dimensional effective description} suffices to predict its behaviour. They draw a critical distinction between two regimes: \textbf{Knowledge-Out (KO)} emergence, where complex macroscopic structure arises from simple homogeneous components following local rules (as in many-body physics), and \textbf{Knowledge-In (KI)} emergence, where structure arises from systems trained on or adapted to complex inputs---precisely the regime LLMs occupy. The KI label raises the evidential bar: when each component is trained on individualised data, distinguishing genuine emergence from sophisticated engineering requires evidence of \emph{internal coarse-graining}, not only macroscopic behaviour. They propose five (non-exclusive) conditions for emergence: (i) \emph{scaling} yielding a new parsimonious description; (ii) \emph{criticality} or phase transitions in a control variable; (iii) \emph{compression} of internal representations exploited by the model; (iv) discovery of \emph{novel bases} forming an internal ``alphabet''; and (v) \emph{generalisation} across qualitatively different tasks. This framework is normative rather than purely descriptive: it explicitly challenges most of the existing LLM emergence literature to verify, for each putative emergent ability, the existence of a corresponding internal mechanism rather than relying on benchmark accuracy curves alone.

The definitions above disagree less about what emergence \emph{is} than about where to look for it.
The field's central dispute over whether the capability jumps of Wei et al.~\citep{wei2022emergent} are
real or a metric artifact~\citep{schaeffer2024emergent} plays out entirely on benchmark curves. Yet
an abrupt curve is neither necessary nor sufficient for emergence in Anderson's~\citep{anderson1972more}
sense. What matters is whether a new, lower-dimensional description of the network acquires its own
effective laws that screen off the underlying weights. 

What matters is whether a new, lower-dimensional description of the network acquires its own effective laws that screen off the underlying weights. We therefore take as an evaluative lens, following Krakauer et al.~\citep{krakauer2025complex}, a criterion that is mechanistic rather than purely behavioral: an ability is most convincingly emergent when a compressed, causally efficacious internal mechanism explains it. This bar is high for LLMs specifically, since in their Knowledge-In regime an impressive capability may reflect sophisticated data interpolation rather than genuine reorganization. Rather than re-adjudicating every reported ability against this standard, we use it where the evidence allows to ask (1) whether a low-dimensional circuit tracks a given ability, (2) whether intervening on it changes the behavior, and (3) whether it screens off the weights; a growing body of interpretability work~\citep{yang2025emergent, li2025grokking, zucchet2025sparse} is beginning to supply such evidence for a handful of abilities, while most of the 137 catalogued by Wei et al.\ remain untested against it (\cref{sec:mirage,subsec:arith}).

\begin{table*}[!h]
    \centering
    \caption{Summary of the sections and subsections in this survey. Each row provides an overview of the topic and a brief summary.}
    \label{tab:subsections_summary}
    \fontsize{9pt}{10.5pt}\selectfont{
    \begin{tabular}{p{4.5cm}p{8.5cm}}
        
        \toprule
        \textbf{Topics} & \textbf{Findings Summary} \\ \midrule
        
        \textit{Section 3: Emergent Abilities in Large Language Models} & 
        Reviews evidence from benchmarks showing abrupt, task-specific improvements as models scale up, highlighting the inherent unpredictability of such emergent behavior. \\ \midrule

        \textit{Section 3.1: Do emergent abilities really exist?} & 
        Examines the mirage debate, showing that while continuous metrics smooth some jumps, discontinuities persist as bimodal seed distributions, population-scale gradual curves, and internal coarse-grained mechanistic reorganizations. \\ \midrule
        
        \textit{Section 3.2: Relationship between Emergent Abilities and Prompt Strategies} & 
        Explores how few-shot prompting, CoT prompting, and instruction tuning trigger emergent-like behavior on multi-step reasoning tasks, linked to representation superposition. \\ \midrule
        
        \textit{Section 3.3: Loss Functions and Emergent Abilities} & 
        Demonstrates that pre-training loss predicts downstream capabilities via domain capacity allocation, though this relationship can completely decouple under out-of-distribution shifts. \\ \midrule
        
        \textit{Section 3.4: The Impact of Quantization on Emergent Abilities} & 
        Investigates how compression triggers non-linear Phase Transition Points where performance collapses, noting that larger models delay this via size-dependent redundancy buffers. \\ \midrule
        
        \textit{Section 3.5: Emergent Abilities and Task Complexity} & 
        Explores how overlapping U-shaped and inverted-U scaling patterns across varying task difficulties can create a macro-statistical illusion of stagnant performance. \\ \midrule

        \textit{Section 3.6: Emergent Internal Representations and Mechanisms} & 
        Discusses the spontaneous formation of structured internal computations, including arithmetic state representations, three-stage symbolic circuits, and language-agnostic neurons. \\ \midrule

        \textit{Section 3.7: Predicting Emergent Abilities} & 
        Summarizes forecasting methods like PASSUNTIL, FLP loss mapping, proxy tasks, and parametric laws, alongside routing-consistency monitors for pre-training grokking dynamics. \\ 
        \midrule
        \midrule
        \textit{Section 4: Emergent Abilities as In-context Learning} & 
        Analyzes ICL as a scale-dependent mapping capability driven by sparse attention feedback loops, noting its vulnerability to data shifts and its transience due to strategy coopetition. \\ 
        \midrule
        \midrule
         \textit{Section 5: Emergent Abilities of Large Reasoning Models} & 
        Examines how LRMs execute multi-step tasks via scaled RL and test-time search, leveraging exploration rewards to elicit a hierarchical shift from procedural to strategic planning. \\ \midrule
        \midrule
        \textit{Section 6: Emergent Behaviors in LLMs-powered AI agents} & 
        Discusses agentic multi-step execution and details group-level multi-agent dynamics, where spontaneous conventions, minority-driven shifts, and population-scale alignment emerge. \\ 
        \midrule
        \midrule
        \textit{Section 7: Emergent Harmful Behaviors in LLMs and LLMs-powered AI agents} & 
        Details how models develop implicit deception, manipulation, and reward hacking, showing that minor data contamination or narrow tuning induces broad, transferable misalignment. \\ \midrule

        \textit{Section 7.1: AI Safety} & 
        Addresses governance and regulatory challenges for highly autonomous systems, analyzing global policy fragmentation (e.g., EU AI Act) and the critical need for continuous oversight frameworks. \\ \midrule

    \end{tabular}}
\end{table*}

\section{Emergent Abilities in Large Language Models}\label{sec:emergent_abilities_llms}

Emergent abilities in LLMs have been a subject of increasing interest, particularly since their characterization in~\citep{wei2022emergent}. The phenomenon can be understood through an analogy to phase transitions in physics, where a system undergoes a sudden qualitative shift in behavior once a critical threshold is crossed. In the case of LLMs, these emergent abilities are not a product of gradual improvement but instead appear abruptly when scaling reaches a certain level. Performance often hovers near random until this threshold is surpassed, at which point a sharp jump occurs. 

This unpredictability makes it difficult to foresee emergent abilities by simply extrapolating from smaller models. \textit{It also raises fundamental questions about the complexity and non-linear scaling behavior of LLMs}. 
For this section, we have selected all the papers that appeared when searching in Google Scholar with the query \textit{"Emergent Abilities" "Large Language model"}.


A key early investigation into emergent abilities was conducted through the BIG-Bench benchmark \citep{srivastava2022beyond}. This study proposed two key indicators for emergent behaviors: \textit{linearity} and \textit{breakthroughness}. The authors observed that certain tasks, such as figure-of-speech detection, periodic element identification, and modifier arithmetic, exhibited high breakthroughness, meaning their performance jumped unpredictably at a certain scale. Interestingly, they noted that when using  ``smoother'' evaluation metrics that allow for partial credit, these abrupt leaps disappeared. This raises an important question: \textit{if an ability appears emergent under a binary metric, but follows a continuous trajectory under a more fine-grained metric, should it still be considered emergent?} Indeed, Elhady et al.~\cite{elhady2025cpt} provide concrete evidence that the loss-emergence relationship can break down entirely under distribution shift, with two models matching in perplexity yet differing dramatically in emergent downstream behavior, reinforcing the need for the mechanistic perspective we advocate below.

Ganguli et al.~\citep{ganguli2022predictability}  further investigated this phenomenon and found that while overall test loss decreases smoothly and predictably with increased model size and training data, individual NLP tasks often show abrupt, non-linear improvements. For instance, while perplexity, the standard measure of language model fluency, declines in a steady manner with scale, certain tasks, such as three-digit addition and program synthesis, demonstrate sudden jumps. A striking example is three-digit addition accuracy: a model with 6B parameters achieves only 1\% accuracy, a 13B model improves slightly to 8\%, but a 175B model suddenly reaches 80\% accuracy. This stark contrast between smooth loss improvement and sudden task-specific breakthroughs highlights the enigmatic nature of LLM scaling.

A broader analysis of emergent abilities was conducted by Wei et al.~\cite{wei2022emergent}, who examined multiple LLMs -- GPT-3 \citep{brown2020language}, LaMDA \citep{thoppilan2022lamda}, Gopher \citep{rae2022scaling}, PaLM \citep{chowdhery2022palm}, Chinchilla \citep{hoffmann2022training} --  across different scaling metrics, such as model size and training computation. They categorized tasks into few-shot prompted tasks and those benefiting from augmented prompting techniques like chain-of-thought reasoning and fine-tuning. Their findings reinforced the idea that emergent behaviors are not only unpredictable but also uncapped in scope; LLMs may develop new, unforeseen capabilities, including potentially harmful ones \citep{perez2022ignorepreviouspromptattack, bai2022training, hagendorff2024deception, williams2024targeted}. Generally, the previous works argue that there are no clear trends for the types of tasks that are most emergent. There is an exception in \citep{wei2022emergent}, where Wei et al. analyzed the MMLU benchmark and showed how social Science and Humanities were the most emergent subjects. 
Expanding on BIG-Bench's observations, Wei et al.~\citep{wei2022emergent} further analyzed that emergent behaviors might be, at least in part, an artifact of metric selection. Since most evaluations rely on accuracy-based (binary) metrics that do not award partial credit, performance jumps can appear more sudden than they might actually be. To test this, they compared cross-entropy loss, evaluated on the test set of the specific task, against traditional error rate evaluations and found that loss generally improves smoothly, even when accuracy seems to exhibit an abrupt transition. However, exceptions exist; for instance, in module arithmetic, periodic elements, French-English translation, and IPA transliteration tasks, partial credit metrics exhibited a sharp performance jump, reinforcing that some abilities may be truly emergent, regardless of the evaluation metric used.
Further evidence for emergent abilities with continuous metrics was presented by Steinhardt et al.~\citep{steinhardt2022emergent}, who identified a sudden jump in performance for the French-to-English translation task (WMT-14 Fr-En) when measured using the BLEU score.

For a comprehensive list of \textbf{137 identified emergent abilities}, we refer the reader to the Appendix of \citep{wei2022emergent}.

\subsection{Do emergent abilities really exist?}\label{sec:mirage}
\begin{figure}[!t]
    \centering
    \includegraphics[width=\linewidth]{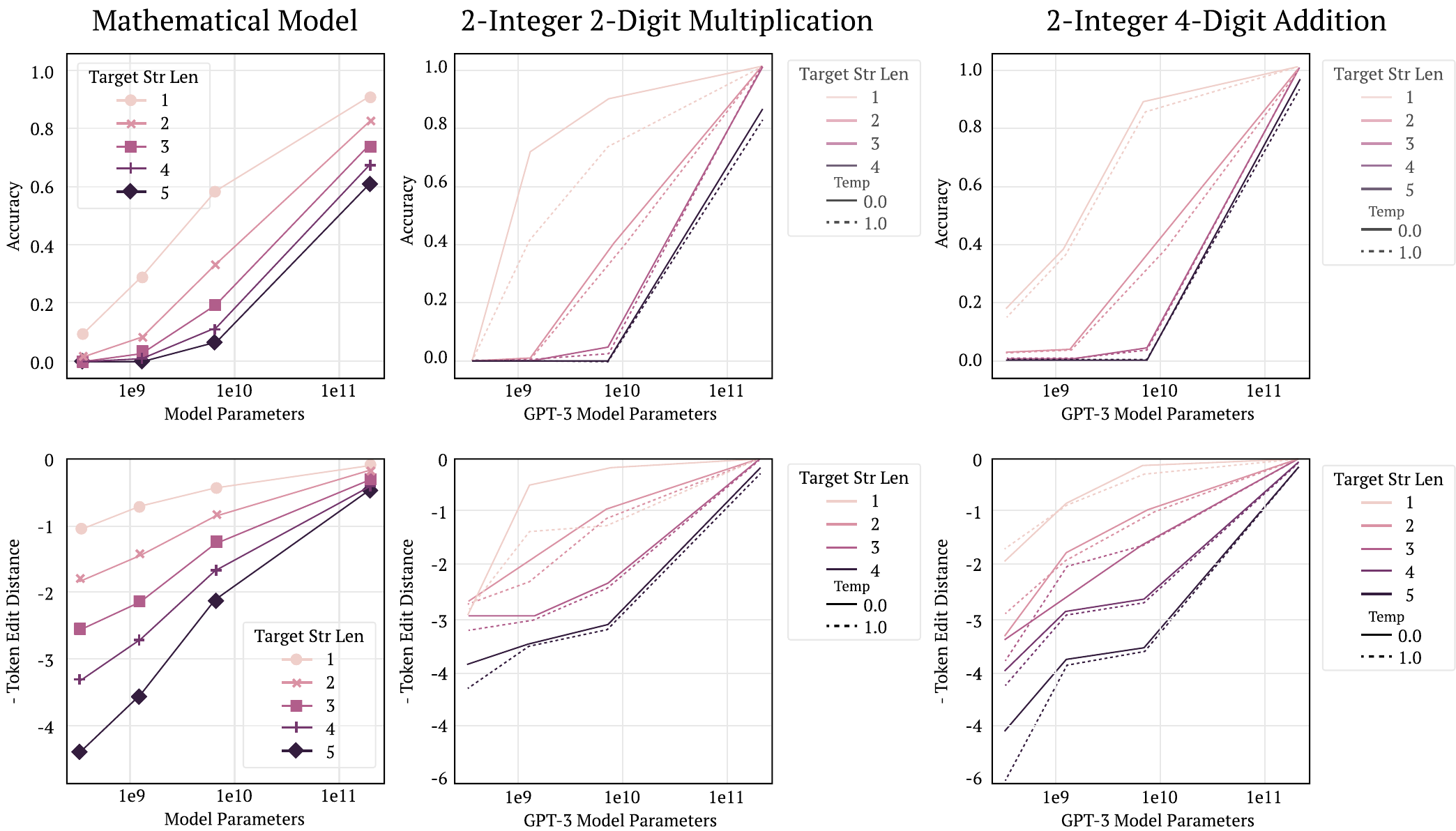}
    \caption{\textbf{Reproduced from \cite{schaeffer2024emergent} (standard deviations on the curves could not be reproduced due to missing data).} \\
    Original caption: \textit{\textbf{Claimed emergent abilities evaporate upon changing the metric.} Left to Right: Mathematical Model, 2-Integer 2-Digit Multiplication Task, 2-Integer 4-Digit Addition Task. Top: When performance is measured by a nonlinear metric (e.g., Accuracy), the InstructGPT/GPT-3 \citep{brown2020language} family's performance appears sharp and unpredictable on longer target lengths. Bottom: When performance is instead measured by a linear metric (e.g., Token Edit Distance), the family exhibits smooth, predictable performance improvements.}}
    \label{fig:shaeffer_fig3}
\end{figure}

The debate over the existence of emergent abilities in LLMs continues, with Schaeffer et al.~\cite{schaeffer2024emergent} extending \citep{srivastava2022beyond, wei2022emergent} to further investigate the role of evaluation metrics in detecting these phenomena. They challenge the notion of sharp, unpredictable capability leaps, arguing that such emergent abilities stem from nonlinear metrics like Accuracy, which prior studies favored. By adopting Token Edit Distance, a metric that awards partial credit, they assert that performance improvements in GPT-3 across arithmetic tasks (e.g., 2-integer 2-digit multiplication, 2-integer 4-digit addition) appear ``smooth, predictable'' rather than abrupt (Fig. 3 of \citep{schaeffer2024emergent}, reproduced in Fig. (\ref{fig:shaeffer_fig3})).
\begin{quote}
``When performance is instead measured by a \hl{linear metric} (e.g., Token Edit Distance), the family exhibits \hl{smooth, predictable performance improvements}.''\\\\\hspace*{\fill} -- Schaeffer et al.
\end{quote}

However, this conclusion is less robust than it seems. For instance, in 2-integer 2-digit multiplication (target lengths 1, 3, 4) and 2-integer 4-digit addition (lengths 4, 5), performance trends exhibit irregularities rather than seamless progression. More critically, Token Edit Distance's suitability for arithmetic proficiency is questionable. Consider the sum \( 4237 + 5487 = 9724 \): an LLM outputting \( 2724 \) incurs just a one-token edit (9 \(\to\) 2), despite a 7000-unit error. This suggests the metric prioritizes syntactic similarity over semantic accuracy, casting doubt on its ability to reflect reasoning skills.

\begin{figure}[!t]
    \centering
    \includegraphics[width=\linewidth]{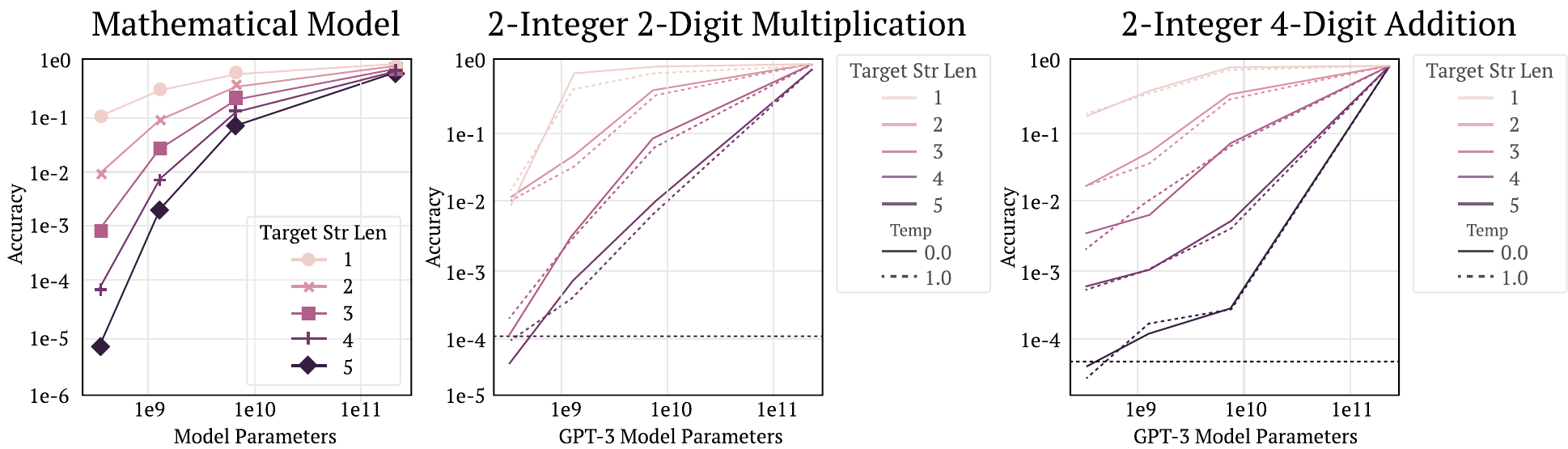}
    \caption{\textbf{Reproduced from \cite{schaeffer2024emergent} (standard deviations on the curves could not be reproduced due to missing data).} \\
    Original caption: \textit{\textbf{Claimed emergent abilities evaporate upon using better statistics.} Left to Right: Mathematical Model, 2-Integer 2-Digit Multiplication Task, 2-Integer 4-Digit Addition Task. Based on the predictable effect Accuracy has on performance, measuring performance requires high resolution. Generating additional test data increases the resolution and reveals that even on Accuracy, the InstructGPT/GPT-3 family's \citep{brown2020language} performance is above chance and improves in a smooth, continuous, predictable manner that qualitatively matches the mathematical model.}}
    \label{fig:shaeffer_fig4}
\end{figure}

Schaeffer et al. further hypothesize that increasing test data smooths performances, eliminating apparent emergence (Fig. 4 reproduced in~\cref{fig:shaeffer_fig4}). Yet, their plots raise concerns. Switching from a linear y-axis (Fig. 3 reproduced in~\cref{fig:shaeffer_fig3}) to logarithmic (Fig. 4 reproduced in \cref{fig:shaeffer_fig4}) can create an illusion of smoothness, thereby obscuring residual jumps. For instance, for 2-digit multiplication and 4-digit addition for lengths 3, 4, and 5, a transition going from $1\mathrm{e}{9}$ and $1\mathrm{e}{10}$ parameters to $1\mathrm{e}{11}$ results in the accuracy increasing from $< 1\mathrm{e}{-1}$ to $\approx 1\mathrm{e}{0}$. Does increasing from 10\% to 100\% not represent a significant jump in performance? There is no "evaporation of claimed emergent abilities".

Moreover, their claim that performance predictability hinges on metric choice falters in tasks like module arithmetic, periodic elements, French-English translation, and IPA transliteration tasks, which reveal discontinuities even under continuous metrics \cite{wei2022emergent}.
Furthermore, they claim that their mathematical model \emph{"qualitatively match"} (see Figure~\ref{fig:shaeffer_fig4}) observed trends, but this statement lacks quantitative rigor (e.g., no error bounds or statistical tests). While Schaeffer et al. highlight metric influence, these inconsistencies and alternative findings call for a more nuanced exploration of emergence in LLMs.

In addition to these experiments, the paper also presents a meta-analysis of emergent abilities within the BIG-Bench framework, evaluating multiple model families and metrics. Their findings suggest that most metrics used in BIG-Bench do not exhibit emergent abilities, although their criteria for classifying a metric-task-model triplet as emergent could be debated. Specifically, the authors define a triplet as emergent when its emergence score exceeds 100, a notably stringent threshold. This score reflects a scenario wherein, across a continuum of models of increasing scale evaluated against a specific metric-task pair, the difference between the maximum and minimum metric values is 100 times greater than the median difference. We contend, however, that an emergence score ranging between 20 and 100 might also reasonably qualify as emergent. For instance, consider four models of progressively larger sizes yielding performance scores of 5\%, 10\%, 20\%, and 90\% on a selected metric-task pair; this configuration produces an emergence score of 8.5, which, \emph{while below the authors’ threshold, still suggests a discontinuous jump in performance}. Additionally, they demonstrated that changing evaluation metrics eliminated signs of emergent behavior for three\footnote{swahili-english-proverbs was not emergent also with the first metric.} tasks within the LaMDA model family. Finally, they induced emergent-like behavior in deep networks across various vision tasks, suggesting that \textbf{emergent abilities can, in some cases, be artificially introduced through experimental design}.

Building upon the exploration of the connection between continuous metrics and emergent capabilities, Du et al.~\citep{du2024understanding} evaluated three LLMS of different size, 1.5, 6 and 32 billion parameters, on two commonly used benchmarks, MMLU~\citep{hendrycks2020measuring} and C-Eval~\citep{huang2024c}, using two continuous evaluation methods, specifically Brier Score and Correct Choice Probability (CCP). These metrics allow for a more nuanced assessment of a model's confidence and decision-making process rather than simply measuring whether an answer is right or wrong. Their analysis showed that even with these smoother metrics, the performance jumps persisted. This reinforced the argument that \textbf{emergent abilities are not merely an evaluation artifact but reflect actual learning dynamics in the model's development}.

While~\cite{du2024understanding} argue that emergent jumps survive smoother metrics, and ~\cite{schaeffer2024emergent} contend that they evaporate under them, ~\cite{zhao2026randomscalingemergentcapabilities} propose a third position that reframes the debate entirely. Their central claim is that emergent breakthroughs are neither a fixed-threshold capacity unlock nor a metric artifact: they are the visible signature of a bimodal distribution of training outcomes across random seeds. Near the breakthrough scale, independent runs cluster into a ``failure'' mode and a ``success'' mode; a single scaling curve drawn from a fixed seed may therefore look sharply emergent or smoothly linear depending entirely on which cluster each scale happens to sample. The discontinuity is real for each individual run, yet the \emph{probability} of falling into the success cluster changes continuously with scale.

To test this hypothesis without the prohibitive cost of pretraining hundreds of seeds, the authors combine two complementary setups. On synthetic length-generalization tasks (counting and reverse-order addition with index hints) they train 200--250 decoder-only Transformers per scale, varying width and depth independently. On natural language, they partially reinitialize the final attention layer and language-model head of Qwen2.5-0.5B and Qwen2.5-1.5B and continue-pretrain 80 seeds on mixtures of C4 news data and the MMLU auxiliary training set, controlling both model scale and target-task data proportion.

Three findings are particularly relevant for the present debate. First, decomposing the bimodal distribution into a \emph{failure} component and a \emph{success} component reveals that the overall \emph{mode} improves abruptly while the \emph{probability of success} and the \emph{mean accuracy of successful runs} improve smoothly and monotonically; the sharp jump in the conventional scaling curve is the projection of a gradual distributional change onto a coarse summary statistic. Second, and most importantly for the ~\cite{schaeffer2024emergent} critique, the bimodality \textbf{does not disappear under continuous metrics}: using a Lipschitz-continuous per-token loss score, KDE estimates of the MMLU NLL distribution remain visibly multimodal, and Hartigan's Dip Test confirms statistically significant deviations from unimodality ($p < 0.001$ at 10\% and 20\% MMLU mixtures). In fact, continuous metrics sometimes \emph{expose} low-performance clusters that thresholded accuracy conceals. Third, the transition from a unimodal to a bimodal performance distribution is itself abrupt and identifies a \textbf{minimum capacity} threshold that does not necessarily coincide with the scale at which a single-seed curve breaks through.

A complementary critique of the standard emergence narrative comes from ~\cite{sun2025breaking}, who argue that most evidence for sharp capability jumps rests on a handful of flagship models and has rarely been subjected to the population-level statistical scrutiny routine in cognitive science. They re-evaluate two large snapshots of the Open LLM Leaderboard - 1186 models on ARC, HellaSwag, MMLU, Winogrande, GSM8K, and TruthfulQA, and 4569 models on the harder IFEval, BBH, MATH Lvl-5, GPQA, MuSR, and MMLU-Pro - using a pipeline that combines ANOVA and Tukey HSD tests, Generalized Additive Mixed Models (GAMMs) with architecture as a random effect, and t-SNE clustering.

Three findings stand out. First, none of the twelve benchmarks exhibits the canonical sharp transition associated with emergent abilities~\cite{wei2022emergent}: the GAMM partial-effect curves are everywhere significant ($p < 0.001$) but {continuous and non-monotonic}, with U-shaped or wave-like patterns and task-dependent saturation thresholds (around 7B parameters for the easier benchmarks and around 84B for the harder ones), beyond which gains become unpredictable and on IFEval, MuSR, and MATH Lvl-5 even \emph{decline}. The authors term this regime \emph{gradual emergent abilities}. Second, ANOVA tests reveal significant gaps between pretrained and post-trained models but {no significant difference between fine-tuned, instruction-tuned, and RL-tuned variants}, suggesting that the finer distinctions among post-training paradigms do not survive aggregation across a representative population. Third, broad reasoning and knowledge benchmarks (ARC and MMLU in the first dataset, BBH and GPQA in the second) emerge as the most integrative abilities, while mathematical reasoning and truthfulness behave more independently and sometimes show \emph{negative} associations with general reasoning, hinting at genuine capability tensions rather than a single underlying factor.\color{black}

Another piece of evidence that emergent transitions are not merely metric artifacts comes from~\cite{gu2025data}, who study factual knowledge acquisition when a knowledge-dense dataset (synthetic biographies) is mixed with a large web corpus. They identify two phase transitions: one with respect to model size and another with respect to the mixing ratio. Crucially, these transitions persist when measured by validation loss, a continuous metric. This result complements~\cite{zhao2026randomscalingemergentcapabilities}: while Zhao et al. show that bimodality across seeds survives smoother metrics, Gu et al. show that sharp transitions across model scales and data compositions survive them too. Together, they constrain the Schaeffer hypothesis — the existence of sharp transitions is not contingent on choosing thresholded metrics. A related strand of evidence comes from \citep{abramov2025grokking}, who elicit grokking, delayed generalization emerging well after the training loss has converged, in real-world multi-hop reasoning by augmenting the underlying fact set. By moving the phenomenon out of synthetic algorithmic settings, they reinforce the view that the abrupt onset of a capability is a genuine feature of training dynamics rather than an artifact of coarse evaluation.

Beyond the metric-artifact debate, Krakauer et al.~\cite{krakauer2025complex} argue that the entire discussion --- whether jumps evaporate under continuous metrics or persist under them --- misses a deeper point. In their complexity-science reading, an accuracy curve alone (smooth or abrupt) cannot settle whether an ability is emergent, because emergence in the technical sense requires the identification of a \textbf{coarse-grained internal mechanism} --- a lower-dimensional effective description of the network that causally explains the behaviour and screens off the microscopic weights. Absent such mechanistic evidence, ``emergence'' collapses into observer-side surprise; conversely, ability jumps accompanied by identified internal reorganisation (e.g., the symbol-abstraction / induction / retrieval circuit of~\cite{yang2025emergent} or the pathway restructuring of~\cite{li2025grokking}) constitute stronger evidence of genuine emergence than any accuracy curve on its own.

\subsection{Relationship between Emergent Abilities and Prompt Strategies}\label{subsec:prompt}
Recent research has introduced various prompting strategies to enhance language model capabilities. These include tailored prompts, such as \textit{chain-of-thought} (CoT) prompting, which guides multi-step reasoning, and advanced fine-tuning approaches, such as instruction following, which adapts models to specific tasks. Wei et al.~\citep{wei2022emergent} found that certain techniques led to sudden jumps in performance, particularly in large models. For instance, CoT prompting significantly improved performance in math word problems because these problems require step-by-step reasoning, which is exactly the type of thinking CoT induces.

Not just prompting but also fine-tuning strategies have shown emergent effects. Wei et al.~\citep{wei2022emergent} further demonstrated that \textit{instruction tuning} \citep{ouyang2022training} \citep{wei2021finetuned}, where tasks are framed as instructions, and \textit{scratchpad reasoning} \citep{nye2021show}, which predicts intermediate steps, yield substantial performance boosts, but only in large-scale models (100B+ parameters). Lu et al.~\citep{lu2023emergent} explored this phenomenon further, disentangling the effects of few-shot prompting, instruction tuning, and CoT prompting to assess emergent abilities in isolation. They questioned whether instruction-tuned LLMs genuinely develop reasoning abilities or simply perform better due to learned heuristics.

Their experiments, conducted on four model families (GPT-3, T5, LLaMA, and Falcon) across 22 tasks, revealed that without few-shot prompting, these models showed no emergent abilities, performing only marginally better than random guessing, except in two cases: \textit{Hindu Knowledge} (which relies on memory) and \textit{Nonsense Word Grammar} (which tests formal linguistic abilities rather than functional reasoning). They concluded that \textit{in-context learning} (i.e., few-shot prompting) is essential for emergent functional abilities, and while instruction tuning improves general performance, it does not lead to genuine reasoning.

While \citep{lu2023emergent} provided valuable insights at the time, it is important to recognize that the field of LLMs has advanced rapidly since their research. More recent models, such as OpenAI o3-mini, Claude 3.5, Gemini 2.0, and DeepSeek-R1, have achieved remarkable advances, calling into question the relevance of its findings. Emerging abilities studies have consistently shown that larger, better-trained models can exhibit fundamentally different and often unpredictable behaviors. For example, Claude 3.5 Sonnet achieved\textbf{ 96.4\% accuracy on GSM8K} (grade-school math) and \textbf{93.7\% on code generation}, while o3 scored \textbf{87.7\% on PhD-level science questions} (GPAQ Diamond), \textbf{surpassing human experts}.  Furthermore, models such as OpenAI's o3 and DeepSeek-R1 have demonstrated significant progress, which may influence the applicability of earlier findings.
These advancements underscore the rapid evolution of LLMs, highlighting the importance of considering these developments when interpreting earlier research conclusions. 

\subsection{Loss Functions and Emergent Abilities}\label{subsec:loss}

\noindent\textbf{Key Experiments and Findings.} Du et al.~\citep{du2024understanding} provide a novel perspective on emergent abilities in LLMs by analyzing their relationship with pre-training loss. Their work investigates whether certain abilities emerge at specific loss thresholds during training, offering a fresh angle on why LLMs suddenly improve in certain tasks as they scale. To explore this relationship, the authors trained three LLMs of different sizes, 1.5, 6, and 32 billion parameters, and evaluated their performance across twelve diverse downstream tasks at multiple checkpoints throughout training. By monitoring these models over time, they observed two major trends.

First, certain tasks, including MMLU, C-Eval, GSM8K, and GSM8K-Chinese, exhibited a distinct threshold in pre-training loss. Once the loss dropped below a critical value, the models' performance on these tasks abruptly improved, suggesting a sudden emergence of ability. This pattern indicates that emergent abilities are not solely a function of scale but are tied to the training process itself. Rather than improving gradually, performance seemed to remain at near-random levels until the loss threshold was reached, after which the models showed a sharp increase in capabilities.

The second key finding was that pre-training loss acted as a strong predictor of downstream task performance, often independent of the model's size. This suggests that beyond mere scale, a model's actual learning progress, as measured by loss reduction, plays a crucial role in determining when and how it develops certain abilities. \textbf{This finding challenges the idea that emergence is purely a consequence of increasing model parameters and instead highlights the importance of training dynamics}.

To strengthen their conclusions, the authors extended their analysis in two important ways. First, they investigated whether the training dataset size influenced the observed relationship between loss and emergent abilities. By varying the size of the dataset, they assessed whether the same loss thresholds applied when more or less data was available. Second, they tested their hypothesis using the publicly available LLaMA models \citep{touvron2023llama}, allowing them to see whether the observed trends were consistent across different model architectures. Both of these extensions reinforced their initial conclusions, demonstrating that pre-training loss can be a reliable predictor of downstream task performance, regardless of dataset size or specific model design. One limitation of this study is that it only examined models within a relatively narrow range, spanning two orders of magnitude. While this provides useful insights, it leaves open the question of whether the same trends hold for even larger models or whether additional scaling leads to new, unforeseen behaviors.

\noindent\textbf{Data Mixing and Capacity Allocation.} \cite{gu2025data} provides a mechanistic answer to why pre-training loss thresholds appear. They show that when an LLM is trained on a mixture of web-scraped data and a small knowledge-dense subset, knowledge acquisition exhibits sharp phase transitions in both the mixing ratio $r$ and the model size $M$: below a critical mixing ratio $r_{\mathrm{thres}}$, the model memorizes essentially nothing from the knowledge-dense data — even after each fact appears thousands of times — and the required training tokens grow exponentially as $1/r$ increases. Their information-theoretic framework models a sufficiently trained LLM as an optimal bounded-capacity learner that allocates capacity across domains according to their "marginal value" (the loss reduction per unit of allocated capacity), analogous to a fractional knapsack solver. Assuming web-data loss scales as $\mathcal{L}2 = A \cdot M^{-\alpha} + C$, they derive $r{\mathrm{thres}} \sim 1/(p \cdot M^{\alpha+1})$ and $f_{\mathrm{thres}} \sim 1/M^{\alpha+1}$, where $p$ is the per-fact exposure frequency. This recasts emergence as a discontinuous reallocation of bounded capacity: as $M$ or $r$ crosses a threshold, the knowledge-dense domain transitions from "not worth learning" to "worth learning," producing the observed jump. The authors also document the same phase transition on simple reasoning tasks (slope calculation, Max-over-N), where models generalize from a vanishingly small fraction of the input space — suggesting the mechanism extends beyond pure fact memorization.

\noindent\textbf{Memorization vs. Generalization.} Huang et al.~\citep{huang2024unified} offers another perspective on the emergence of abilities, connecting it to the competition between memorization and generalization circuits in neural networks. Their work builds on previous research into grokking, a phenomenon where models initially memorize data before abruptly transitioning to generalization \citep{power2022grokking}. The authors extend this framework to different model sizes and training data volumes, demonstrating that similar dynamics may explain emergent abilities in LLMs. Their findings suggest that \textbf{when a model is heavily tasked with memorization, the development of generalization abilities is delayed}. In other words, the presence of a memorization-heavy task can push back the point at which the model begins to generalize, requiring much larger model sizes or significantly more training time for the transition to occur. Their work aligns with \citep{du2024understanding}, reinforcing the idea that emergent behaviors are not just a byproduct of scale but are deeply tied to the learning dynamics of neural networks. While Huang et al. frame the memorization-generalization trade-off as an internal competition between circuits, a complementary view asks whether the contribution of each mechanism can be measured directly from the pretraining corpus, and whether it depends on the nature of the task.

\noindent\textbf{Distributional Memorization Across Task Types}
A scalable empirical counterpart to the circuit-level account of \citep{huang2024demystifying} is provided by \citep{wang2025generalization}, who introduce {distributional memorization}: the Spearman correlation between an LLM's output probabilities and the frequency of task-relevant n-gram pairs in its pretraining corpus. To capture task-specific dependencies that classical n-gram statistics miss, they build a {task-gram language model} by mining semantically related input-output n-gram pairs from a supervised task dataset and counting their co-occurrences in the full Pile via WIMBD and the $\infty$-gram index. Memorization is then defined as a global statistical property of the (model, corpus, task) triple, not as verbatim recall, sidestepping the prohibitive cost of counterfactual retraining that limited prior work.

Applied across the full Pythia family (14M-12B) on four tasks, WMT translation, TriviaQA, MMLU, and GSM8K, the framework reveals a sharp task-type asymmetry. Knowledge-intensive recall (TriviaQA) exhibits strong and {increasing} distributional memorization with model scale ($\rho > 0.35$ at 12B), and knowledge-intensive MMLU subtasks pattern similarly. In contrast, reasoning-intensive tasks (GSM8K, reasoning-intensive MMLU) and translation show {decreasing} memorization and {increasing} generalization with scale: larger Pythia models produce monotonically more novel n-grams on WMT, directly contradicting the in-distribution finding of ~\citep{merrill-etal-2024-evaluating} that bigger models are less novel. A gradient-based training-influence analysis (\citep{pruthi2020estimating}) corroborates the correlational picture causally, with pretraining documents containing task-gram pairs exerting the largest influence on TriviaQA, moderate on MMLU, and negligible on WMT. The practical upshot is that emergent capabilities of the same model are driven by qualitatively different mechanisms depending on the task: scale buys memorization where the answer lives in the corpus, and generalization where it does not. The framework also yields a concrete intervention --- prompts can be optimized to maximize or minimize their pretraining-corpus n-gram footprint, with measurable gains on TriviaQA from memorization-encouraging prompts and on GSM8K from generalization-encouraging ones --- offering an operational handle that complements the Du et al. and Gu et al. perspectives above and motivates the multilingual decoupling discussed by Elhady et al. immediately below.

\noindent\textbf{Limitations and Future Directions in Loss-based Emergent Abilities.}
While \cite{du2024understanding} presents compelling evidence for a strong correlation between pre-training loss and the emergence of abilities, its analysis remains correlational rather than causal. This raises an important question: \emph{why do certain abilities emerge precisely when pre-training loss crosses a particular threshold?} Understanding this process requires more than just identifying patterns; it calls for an investigation into the underlying mechanisms driving these shifts. 

\textbf{Pre-training loss}, while a useful high-level metric, \textbf{does not inherently explain why certain abilities appear at specific points}. The sudden improvement in performance may be linked to deeper changes within the model's internal representations. One possibility is that at certain loss thresholds, the model undergoes structural shifts in \textbf{how it organizes and generalizes knowledge}. These transitions might correspond to qualitative changes in the patterns the model has learned, shifting from surface-level memorization to deeper reasoning and abstraction. However, without a more detailed analysis of the model's internal workings, this remains speculative.

A promising direction for future research is to establish a clearer causal link between pre-training loss and emergent abilities. This could involve analyzing how changes in loss correspond to specific transformations in the model's neural activations, feature representations, or decision-making pathways. We argue that they probably have a common cause or confounder. By combining insights from mechanistic interpretability \citep{bereska2024mechanistic} with the study of emergent abilities, researchers may be able to move beyond mere observation and develop a deeper, mechanistic understanding of why these phenomena occur.

\textbf{Decoupling Perplexity from Emergent Abilities in Continued Pretraining}
A striking empirical challenge to the Du et al.~\cite{du2024understanding} thesis comes from Elhady et al.~\cite{elhady2025cpt}, who study continued pretraining (CPT) of English-centric LLMs for language adaptation to Basque, Arabic, and Indonesian. They observe that two CPT variants of Llama~2 7B, one trained on target-language data only, the other on a mixture of target-language data plus 20\% English, converge to almost identical validation perplexity in the target language, yet differ by up to seven accuracy points on downstream multiple-choice benchmarks. The decoupling is sharp: in Basque, the English-mixed variant exhibits an emergent-style jump of eight points between training steps 2k and 4k on downstream accuracy, while the pure-target variant never takes off, despite the two models being indistinguishable as pure language models on held-out target-language text. This directly contradicts the prediction, central to Du et al. and Xia et al.~\cite{xia2023training}, that models matching in perplexity should match in downstream emergence.

To diagnose the cause, the authors introduce \textbf{Copain}, a language-agnostic in-context learning benchmark composed of seven symbolic tasks (max/min/median of an integer list, odd-among-even, alphabetical first/last) whose prompts contain no natural language and must be inferred from few-shot demonstrations alone. Copain reveals that CPT without English causes a near-total catastrophic forgetting of in-context learning capabilities within the first 2k training steps, whereas mixing in English preserves over 94\% of the original ICL performance. This ICL collapse occurs long before the gap is visible in downstream accuracy: a perplexity analysis of correct vs. incorrect answer labels shows that the model's ability to discriminate the right choice degrades immediately, but the discrimination gap only becomes wide enough to flip the argmax around step 3k.

Mechanistically, the loss-perplexity decoupling tracks a large parameter shift: the cumulative layer-wise $L_2$ distance from the initial Llama~2 weights is roughly 7× higher at step 100 and 15× higher at step 1000 for the no-English variant, with the no-English model undergoing a larger change in the first ten steps than the English-mixed variant does in the entire training run. Two interventions that constrain this shift recover emergent downstream behavior without any English data: a \emph{curriculum} that includes English only during the first 10\% of training, and an \emph{exponential moving average} of model weights applied as a regularizer. LoRA achieves an even smaller parameter shift but underperforms on downstream tasks, suggesting a non-trivial sweet spot. The broader message for the loss-based account of emergence is that pre-training loss reflects only the in-distribution objective; out-of-distribution emergent behaviors are governed by training-dynamics quantities (parameter shift, ICL preservation) that are invisible to the loss curve. This anticipates and concretely instantiates the mechanistic-causal gap flagged in the limitations of Du et al. discussed below.

\subsection{The Impact of Quantization on Emergent Abilities}\label{subsec:quant}

Liu et al.~\citep{liu2023emergent} investigate a critical but often overlooked factor affecting the emergent abilities of LLMs: i.e., quantization. As LLMs grow in size, their memory and computational requirements become increasingly demanding, prompting the need for quantization techniques that reduce precision and optimize efficiency. However, a key question remains: \emph{Does quantization compromise the emergent abilities that make these models so powerful?} Their study systematically examines this trade-off, particularly in the context of in-context learning, chain-of-thought reasoning, and instruction following -- three hallmarks of advanced LLM capabilities.

\noindent\textbf{How Quantization Affects Performance.}
Quantization is a widely used technique that compresses neural networks by reducing the number of bits used to represent each parameter. While this drastically lowers memory usage and inference costs, it often comes at the expense of model performance. Liu et al.~\citep{liu2023emergent} conduct extensive empirical evaluations on a range of LLaMA models, spanning 7B to 65B parameters, across multiple quantization levels, including \textbf{2-bit, 4-bit, 8-bit, and 16-bit precision}. Their findings offer a nuanced understanding of how quantization impacts emergent abilities. At \textbf{higher bit levels}, such as 8-bit and 16-bit, LLMs retain much of their original performance, with minimal degradation in reasoning and instruction-following tasks. However, as precision decreases, particularly at \textbf{4-bit and below}, a clear pattern emerges: while \textbf{4-bit quantization manages to preserve most emergent abilities, 2-bit quantization severely degrades performance}, often reducing accuracy to near-random levels. This suggests that there exists a critical threshold beyond which the model struggles to maintain the structured reasoning and learning dynamics that enable emergent behaviors.

A particularly insightful aspect of the study is its component-wise analysis, which reveals that \textbf{not all parts of the model are equally affected by quantization}. The feed-forward networks within transformer blocks are found to be especially crucial for retaining performance. When these layers undergo aggressive quantization, the model experiences significant losses in reasoning ability and generalization. This highlights the fact that while weight compression is beneficial, indiscriminate quantization of all components can be detrimental.

\noindent\textbf{Can Emergent Abilities Be Preserved?} 
Despite these challenges, the study identifies \textbf{promising post-quantization recovery techniques} that can help mitigate performance degradation. Liu et al. find that fine-tuning after quantization, particularly using parameter-efficient adaptation methods such as LoRA~\citep{DBLP:conf/iclr/HuSWALWWC22}, can significantly restore performance in low-bit models. This suggests that while extreme quantization disrupts emergent abilities, \textbf{targeted fine-tuning can help models adapt and recover key capabilities}. Their findings carry important implications for the future of efficient LLM deployment. \textbf{Low-bit quantization is feasible}, but careful consideration must be given to which model components are quantized and how performance can be recovered. As demand grows for deploying powerful LLMs on resource-constrained devices, understanding these trade-offs will be essential.

These insights open the door for future research into adaptive quantization techniques, where different components of a model are quantized at varying levels based on their importance to reasoning and learning. Additionally, \textbf{quantization-aware training}, where models are trained with low-bit precision from the start rather than applying quantization post-hoc, might further mitigate performance loss.

Ultimately, as LLMs continue to scale, the ability to \textbf{balance efficiency and emergent intelligence} will be crucial. Optimizing quantization strategies without sacrificing core reasoning and learning abilities will not only make these models more practical but also ensure that they remain powerful tools for a broad range of real-world applications.

\noindent\textbf{From Quantization Thresholds to Model Phase Transitions.} Ma et al.~\cite{ma2026phase}
generalise this threshold behaviour beyond quantization into a unifying framework of \textbf{model
phase transitions}: across pruning, quantization, and low-rank decomposition, performance degrades
gracefully up to a critical \emph{phase transition point} (PTP), beyond which it collapses abruptly
and non-linearly. They attribute this to three \emph{orthogonal} forms of redundancy---structural
(exploited by pruning), numerical (by quantization), and algebraic/low-rank (by decomposition)---whose
induced errors are additive rather than multiplicative, so that combining methods leaves each method's
PTP essentially unchanged. Benchmarking roughly 40 methods on LLaMA-2, Qwen-2.5, and Gemma-3, they
locate consistent thresholds: unstructured pruning collapses around $55$--$65\%$ sparsity, structured
pruning around $30$--$45\%$, quantization below $3$-bit, and low-rank decomposition around
$20$--$30\%$, giving a robustness ordering of numerical $>$ structural $>$ algebraic. Crucially for
emergence, the PTP behaves like an \textbf{``event horizon of capability''}: beyond it a model does
not merely weaken but qualitatively collapses, losing the emergent abilities that define LLMs---and,
echoing the scale-dependence of emergence, larger models exhibit \emph{delayed} PTPs (a 70B model
retains $94\%$ of its baseline perplexity and $90\%$ MMLU accuracy even at 2-bit, whereas sub-10B
models lose $\geq 30\%$ perplexity and $25\%$ accuracy), which the authors attribute to
size-dependent ``redundancy buffers''. Exploiting the orthogonality, their criticality-aware
``phase-avoidance'' strategy compresses a model to roughly $10\%$ of its size near-losslessly, and a
compressed LLaMA-2-7B outperforms a natively trained LLaMA-3.2-1B of comparable footprint
(ARC-Challenge $55.0$ vs.\ $45.0$), suggesting that compressing a large model preserves emergent
``world-model'' features that a small model never acquires.

\subsection{Emergent Abilities and Task Complexity}\label{subsec:complex}
\begin{figure}
    \centering
    \includegraphics[width=1.0\linewidth]{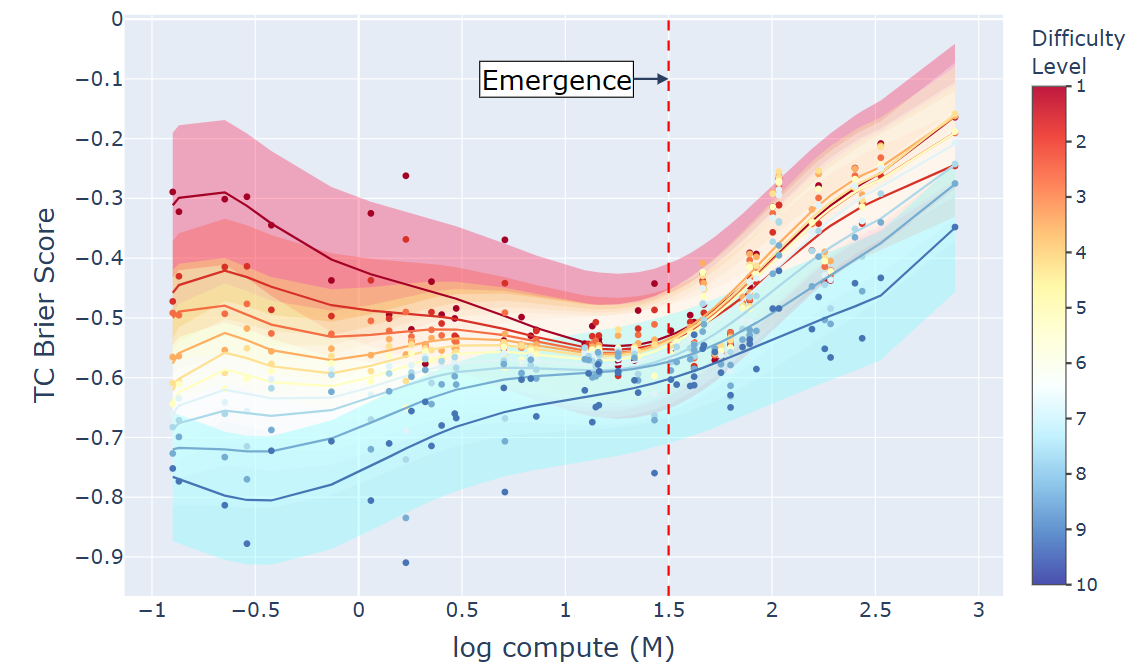}
    \label{fig:enter-label}
    \vspace{-0.1cm}
    \caption{Reproduced from \cite{wu2024u}. Original Caption: U-Shaped and inverted-U scaling with MMLU’s questions clustered into 10 groups. Higher levels are harder questions.}
\end{figure}
The prevailing narrative surrounding emergent abilities in large language models has long emphasized model scale as the primary driver. The assumption was straightforward: as models grow larger and are trained on more data, they suddenly exhibit new capabilities. However, \citep{wu2024u} offer a compelling counterpoint, directing attention to task complexity as a crucial, and perhaps a previously underappreciated, factor in the emergence phenomenon. Wu et al.~\citep{wu2024u} take a new approach by analyzing how \textbf{performance scaling patterns vary across tasks of different difficulty levels}. Their research categorizes questions within downstream tasks by difficulty level, uncovering two distinct trends that interact in unexpected ways:
\begin{enumerate}[topsep=0pt,noitemsep]
    \item \textbf{U-shaped scaling for harder questions} -- Initially, performance declines as models scale before improving after crossing a certain threshold.
    \item \textbf{Inverted-U scaling for easier questions} -- Performance peaks early, then temporarily declines, before improving again as models scale further.
\end{enumerate}%
We reproduce the main figure in \ref{fig:enter-label}. At first, these opposing patterns cancel each other out, leading to the illusion of stagnant performance. However, once models cross a critical scale, the \textbf{reversal of these trends triggers a sudden leap in performance}, giving rise to emergent abilities. This finding suggests that emergence is not necessarily about acquiring new capabilities out of nowhere but rather about overcoming a hidden trade-off between task difficulty and model capacity.

To predict and explain these emergent behaviors, Wu et al.~\citep{wu2024u} introduce the Slice-and-Sandwich pipeline. This method categorizes tasks by difficulty, fits polynomial regression curves separately to easy and hard questions, and then forecasts performance beyond the threshold where emergent behaviors appear. They apply this technique to MMLU \citep{hendrycks2020measuring}, Persian-QA \citep{PersianQA}, and arithmetic benchmarks, demonstrating that it outperforms traditional sigmoid-based scaling laws in predicting when and how emergent abilities manifest.

\noindent\textbf{Implications and Limitations.}
Wu et al.~\citep{wu2024u} challenges the conventional wisdom that emergent abilities are purely a result of scale. Their findings indicate that the \textbf{illusion of stagnation in performance may simply be a result of competing scaling trends} rather than a fundamental limit of the model. However, there are some limitations. For example, the authors focus primarily on multiple-choice tasks, meaning their scaling trends may not generalize to open-ended generative tasks, which rely on different cognitive and linguistic skills. Additionally, the Slice-and-Sandwich pipeline relies on retrospective data to identify difficulty levels and emergence thresholds, which may limit its predictive power in unseen tasks.

\subsection{Emergent Internal Representations and Mechanisms}\label{subsec:arith}
The study by Chen et al. \citep{chen2024states} investigates how large language models (LLMs) solve multi-digit arithmetic problems, revealing that in those models, Implicit Discrete State Representations (IDSRs) emerge in their hidden states. The research challenges traditional views on arithmetic processing in LLMs, showing that instead of just memorizing facts or manipulating tokens, LLMs create IDSRs through a digit-by-digit process similar to human calculations. These representations evolve across model layers, with a critical transition around layer 10, suggesting that IDSRs act as intermediate storage for partial results. While IDSRs improve arithmetic accuracy, handling longer sequences is problematic, possibly due to information loss. The findings indicate that enhancing IDSRs could improve LLMs' reasoning in diverse mathematical and complex tasks, potentially leading to more reliable and interpretable models.
Going beyond arithmetic, Yang et al.~\cite{yang2025emergent} provide a mechanistic account of how abstract reasoning itself is implemented inside LLMs. Combining causal mediation analysis, attention pattern analysis, representational similarity analysis, and ablation studies, they identify a consistent three-stage architecture: (1) symbol abstraction heads in early layers convert input tokens into abstract variables based on their in-context relations, with value embeddings that are invariant to token identity; (2) symbolic induction heads in intermediate layers perform sequence induction over these variables rather than over literal tokens; (3) retrieval heads in later layers convert the predicted variable back into a token, implementing a form of indirection in which variables act as pointers. A linear decoder recovers the abstract variable (e.g., A vs.\ B in an ABA-vs-ABB rule induction task) with >98\% accuracy on completely disjoint token sets, directly evidencing the invariance of these representations.

The architecture replicates across three reasoning tasks (algebraic rule induction, letter string analogies, verbal analogies) and three model families (Gemma-2, Qwen2.5, Llama-3.1) at scales from 2B to 72B parameters. Notably, the four GPT-2 variants (124M--1.5B) lack robust symbol abstraction heads and perform poorly on the rule induction task — suggesting that symbolic mechanisms themselves are an emergent capability that only appears beyond a certain scale or training-data threshold. This finding pushes back directly against characterizations of LLMs as stochastic parrots'' or approximate retrieval'' engines: rather than statistically approximating their training data, LLMs that solve abstract reasoning tasks do so by developing genuine, structured symbolic computations.

\noindent\textbf{Emergent Planning Representations.} Dong et al.~\cite{dong2025planning} probe LLM prompt representations, the hidden states produced \emph{before any response token is generated}, and find they already encode global attributes of the entire upcoming response, including response length, reasoning steps, character choices in storytelling, multiple-choice answers, answer confidence, and factual consistency. Across six model families (Llama-2/3, Mistral, Qwen-2/2.5; 1.5B--72B), simple MLP probes recover these attributes with non-trivial accuracy, revealing a layer-wise hierarchy in which behavior attributes are encoded early, structure attributes peak in mid-layers, and content attributes consolidate late. Unlike abrupt scale-threshold emergence, planning representations are detectable even in smaller models (1.5B), with only weak, family-specific improvements as size grows; the phenomenon is thus better understood as an \textbf{implicit capability that arises from next-token pre-training} rather than as an ability gated by model scale. Probe-based predictions systematically beat the model's own verbalized self-estimates, suggesting LLMs encode richer planning information internally than they can explicitly access during generation.

\noindent\textbf{Emergent Language-Agnostic Representations.} Chen et al.~\cite{chen2025abstract}
ask whether LLMs genuinely ``think in English'' or instead operate in a higher-level space
beyond any single language. Analysing 20 open-source models (Llama, Qwen, Gemma) across six
typologically diverse languages, they identify \emph{language-related neurons}---fewer than
1\% of all neurons---and split them into \textbf{language-shared} (consistently activated
across all languages) and \textbf{language-exclusive} (specific to one language). Two trends
emerge across successive model generations: the proportion of shared neurons rises, and, more
tellingly, deactivating shared neurons degrades performance disproportionately more than
deactivating exclusive ones, indicating that shared neurons evolve into \textbf{language-agnostic
neurons} that support abstraction and reasoning rather than mere multilingual processing. They
summarise this with a \emph{Language-Agnostic Score} that grows with model release date and
correlates with multilingual ability, framing the formation of this compact language-agnostic
parameter space as the emergence of \textbf{abstract thought not tied to any linguistic system}.
As with the planning and symbolic-mechanism findings above, the effect tracks model development
rather than raw parameter count, and it motivates neuron-targeted training: models lacking
language-agnostic neurons benefit from training any language-related neuron, whereas in models
where abstract thought has emerged, further gains come from training the language-exclusive
neurons that capture language-specific nuance.

\noindent\textbf{Emergent Superposition in Continuous Chain-of-Thought.} A continuous analogue
of these emergent internal mechanisms is studied by Zhu et al.~\cite{zhu2025superposition} for
\emph{chain of continuous thought} (COCONUT), where a model's reasoning trace is kept in a
continuous latent space rather than decoded to discrete tokens. Prior theory established that a
two-layer transformer with continuous CoT can solve directed-graph reachability by
\textbf{superposition}---holding multiple plausible reasoning traces in parallel within a single
continuous thought---but left open whether gradient training actually discovers this construction.
Analysing the training dynamics in two stages (autoregressive thought generation and final-answer
prediction), the authors show that it does: the \emph{index-matching logit}, which measures the
model's local-search strength, first grows and then stays \textbf{bounded}, in contrast to
discrete-CoT settings where analogous logits diverge. A bounded logit balances exploitation
(following locally promising edges) against exploration (assigning comparable weight to several
traces when the answer is uncertain), and this balance is precisely what yields superposition---even
when each training example provides only a single demonstration path. Experiments on a two-layer
GPT-2-style transformer trained on graph reachability (ProsQA) confirm the predicted logit
trajectories and reveal \textbf{length generalization}: once superposition emerges in early training
stages, the model reuses it to expand its search frontier over more reasoning steps than it was ever
explicitly trained on. The analysis is confined to a simplified two-layer model on a synthetic
reachability task, but it gives a mechanistic account of how implicit parallel reasoning can emerge
from standard gradient-based training.

\textbf{Emergent World Knowledge as a Reconstructed Collective Model.} While the accounts above characterize which internal mechanisms emerge, \cite{taniguchi2026generative} address why a disembodied model acquires world-congruent representations, proposing the Collective World Model hypothesis: human language externalizes a collective world model that an LLM learns only as a statistical approximation. They separate a subjective Type~1 world model, learned by an embodied agent through sensorimotor interaction, from an objective Type~2 model of entities and their relations, noting the paradox that LLMs seem to possess the latter without any access to the world. Their Generative EmCom framework, grounded in Collective Predictive Coding, casts language emergence as decentralized Bayesian inference in which $K$ agents minimize a collective free energy, encoding their grounded latent states $\{z^k\}_k$ into a shared symbol system $m$. The result is a society-scale encoder--decoder chain $\{z^k\}_k \to m \to z_{\mathrm{LM}}$: since $m$ is the sole informational bottleneck between the collective human mind and the model, an LLM that models $p(m)$ must reconstruct a latent space $z_{\mathrm{LM}}$ whose relational geometry mirrors that of $\{z^k\}_k$. Distributional semantics (from $word2vec$ analogies to the cross-modal convergence posited by the platonic representation hypothesis) is thereby recast as inherited rather than learned from scratch, reinforcing the case made above against a ``stochastic parrot'' reading of emergent representations.

\subsection{Predicting Emergent Abilities}\label{subsec:predict}
Several recent studies \citep{hu2023predicting, openai2023gpt, schaeffer2024has, zhang2024predictable, chen2024scaling, snell2024predicting} have explored different methodologies for forecasting downstream performance, particularly in tasks where capabilities appear suddenly as models scale. \textbf{While early scaling laws provided some insight, they often fail to anticipate discontinuous leaps in performance}, a defining characteristic of emergent abilities. These studies seek to refine our understanding of predictive modeling, offering new frameworks that range from statistical scaling laws to fine-tuning-based emergence predictions.

\noindent\textbf{High-Resolution Metrics and Scaling Laws.}
Hu et al.~\citep{hu2023predicting} introduce PASSUNTIL, an evaluation metric designed to detect even the smallest performance improvements during model scaling. Unlike traditional discrete accuracy metrics, PASSUNTIL estimates the probability of success through extensive random sampling, providing a theoretically infinite resolution. This finer granularity enables more precise detection of emergent behaviors, allowing the authors to derive a task-scaling law that enhances the predictability of performance growth in larger models based on observations from smaller ones. Additionally, they propose a hypothesis grounded in multiple neural circuits \citep{elhage2021mathematical}, suggesting that \textbf{emergent behaviors arise when specialized circuits activate at certain scaling thresholds}. However, this study is limited to models up to 2.4 billion parameters, meaning its findings may not be generalized to larger models, where emergence is most prominent.

The GPT-4 technical report \citep{openai2023gpt} also tackles predictability but from a resource efficiency perspective. It suggests that GPT-4's performance can be anticipated using less than 1/10,000th of its full computational resources. However, the methodology behind these predictions remains undisclosed, and the report acknowledges that certain emergent abilities remain unpredictable.

\noindent\textbf{Why Is Predicting Emergence So Difficult?}
Schaeffer et al.~\citep{schaeffer2024has} attempts to explain why downstream performance prediction remains a challenge. Their study, conducted across five model families and twelve diverse multiple-choice benchmarks, identifies a series of statistical transformations that occur when converting negative log-likelihood scores into task performance. These transformations gradually weaken the correlation between model scale and downstream accuracy, making direct prediction increasingly difficult.

Additionally, they investigate how probability mass is allocated to correct vs. incorrect answers, revealing that emergent improvements do not always follow simple trends. This finding suggests that emergence is not merely a function of increased model size but involves complex redistributions of probability across outputs, which standard scaling laws struggle to capture. However, a key limitation of their work is that it focuses exclusively on multiple-choice benchmarks, raising questions about whether these findings extend to open-ended tasks like reasoning, code generation, or summarization.

\noindent\textbf{Proxy Tasks as Predictors.}
Zhang et al.~\citep{zhang2024predictable} introduce a different approach, leveraging proxy tasks that is simpler, smaller-scale evaluations that correlate strongly with later-stage emergent abilities. They analyze datasets such as C3, CMNLI, OCNLI, CHID, RTE, and CMMLU, demonstrating that early performance on these tasks can serve as a reliable predictor of future capabilities in larger models. This structured methodology presents a practical framework for guiding model development and resource allocation, as it allows researchers to anticipate emergent behaviors without requiring full-scale pre-training. Unlike conventional scaling laws, which often fail when applied to unseen capabilities, \textbf{proxy task-based predictions offer an empirical foundation for understanding which skills will emerge at larger scales}.

Complementing these proxy-task methodologies, Ye et al.~\citep{ye2023predictable} directly investigate the predictability of LLMs' capabilities across a wide range of tasks from the BIG-bench benchmark. Using an MLP-based predictor trained on a vast dataset of past LLM experiments, they demonstrate remarkably high predictability in standard train-test splits, achieving R² scores exceeding $95\%$. This suggests that learnable patterns exist within LLM performance data. However, the authors also highlight that predictability diminishes under more challenging conditions, such as when generalizing to completely unseen combinations of model families and tasks. \textbf{Crucially, they find that while emergent abilities are indeed harder to predict than non-emergent ones, they are not entirely unpredictable.} Furthermore, to address the practical challenges of evaluating increasingly complex LLMs, they introduce the concept of a "small-bench", a carefully selected subset of tasks designed to predict performance on the full benchmark efficiently. Their work underscores that while broad patterns of LLM capability scaling are learnable, the nuances of emergent behaviors and generalization across diverse settings remain a significant hurdle for precise prediction.

\noindent\textbf{Predicting Performance with Pre-Training Loss.}
Chen et al.~\cite{chen2024scaling} take a complementary approach, proposing a two-stage prediction framework (FLP) based on pre-training loss. Their method first establishes a function that maps computational resources (FLOPs) to pre-training loss, using smaller models to extrapolate larger-scale trends. They then apply a regression model to correlate pre-training loss with downstream performance, allowing them to predict task outcomes beyond known emergence thresholds. Experimental results show that FLP achieves relative error margins of 5\% and 10\% for 7B and 13B models, respectively, significantly outperforming traditional FLOPs-to-performance extrapolations. The authors further extend this approach with FLP-M, incorporating multiple data sources into the pre-training process to refine predictive accuracy. By using a two-layer neural network to model the non-linear relationship between pre-training loss and downstream abilities, they demonstrate improved forecasting across various benchmarks. 

However, their methodology has limitations. The study primarily focuses on binary cases, such as code vs. text data, limiting its generalizability to more complex multimodal or hierarchical learning scenarios. Additionally, the emphasis on code-mixing ratios means that predictions might not hold for domains with different dataset compositions.

\noindent\textbf{Predictable Phase Transitions from Capacity Allocation.}
An orthogonal predictive framework is proposed by~\cite{gu2025data}, who show that the threshold frequency $f_{\mathrm{thres}}$ above which an individual fact is acquired by a model follows a power law $f_{\mathrm{thres}} \sim M^{-(\alpha+1)}$, where $\alpha$ is the scaling exponent of web-data loss. The prediction is validated on both synthetic biographies and real-world knowledge extracted from Wikipedia, holding across multiple open-source model families. Unlike methods that predict aggregate task performance, this framework predicts which individual pieces of knowledge a model of a given size will acquire — a finer-grained prediction tied directly to data composition rather than scale alone.

\noindent\textbf{Fine-Tuning as an Emergence Predictor.}
Snell et al.~\citep{snell2024predicting} propose a fine-tuning-based method to predict emergent capabilities. Instead of relying on pre-training loss or proxy tasks, they use task-specific fine-tuning on smaller models and observe how this process shifts the emergence threshold. By fitting an emergence law, a parametric function that maps fine-tuned performance to scaling trends, they can anticipate when an untrained, larger model will exhibit the same capabilities. To validate this approach, they test it on benchmarks where emergent abilities have been previously observed. 
Their findings suggest that fine-tuning-based predictions allow them to anticipate emergent abilities up to four times earlier than traditional methods. An unexpected but crucial insight from their study is that pretraining data quality significantly affects emergence timing. Comparing OpenLLaMA V1 and V2, they found that models trained on higher-quality datasets exhibit earlier emergent behaviors, indicating that scaling alone is not the sole driver of capability development.

However, their method also has constraints. It only allows predictions within a 4$\times$ scaling range, meaning it cannot forecast abilities in models orders of magnitudes larger than those observed. Furthermore, \textbf{it remains unclear whether fine-tuning unlocks latent abilities or merely accelerates their natural emergence}, leaving open questions about the precise mechanisms behind capability development.

\noindent\textbf{Pathway-Based Monitors in MoE Pretraining.} Li et al.~\cite{li2025grokking} present the first empirical evidence of \emph{grokking} in LLM pretraining (7B OLMoE), showing that downstream generalization on math, code, commonsense, and domain-specific QA keeps improving long after the pretraining loss has converged. Crucially, this grokking is \textbf{local and asynchronous}: different data groups memorize at different steps, and their generalization leaps on downstream benchmarks occur with delays that scale with data difficulty. To track this hidden memorization-to-generalization transition without held-out evaluation, the authors introduce two metrics defined on the MoE routing structure: the pairwise edit distance between expert sequences across samples (which decreases as pathways converge to shared routes) and a per-sample pathway-consistency score on cross-layer expert transitions (which increases as routing smooths). Both metrics correlate with downstream test accuracy at $|\rho| > 0.97$, substantially outperforming training and validation loss, and are theoretically grounded in a generalization bound via the effective dimension of the routing kernel, offering a \textbf{fine-tuning-free, zero-cost predictor} of emergent generalization that operates entirely on training dynamics.

\section{Emergent abilities as in-context learning}\label{sec:in_context}
The term \textit{emergent} in the context of LLMs is often used to describe capabilities that arise implicitly as models learn language patterns and structures through next-token prediction. These abilities are assessed through \textbf{few-shot or zero-shot prompting}, where models generalize to new tasks without undergoing explicit fine-tuning. This process, known as \textbf{in-context learning (ICL)}, allows LLMs to infer new patterns and concepts solely from contextual information provided in the prompt.

Unlike Wei et al.'s \citep{wei2022emergent} definition of emergent abilities, which emphasizes sudden performance jumps with increased model scale, in-context learning does not necessarily require abrupt improvements. Instead, it refers to the gradual development of capabilities that enable LLMs to perform tasks for which they have not been explicitly trained. Recent work has further shown that ICL itself can be non-monotonic over training, emerging at intermediate stages and then receding as alternative strategies take over~\cite{singh2023transient,singh2025strategy}, a dynamic we discuss in detail below. The research in this area primarily seeks to understand why LLMs generalize to new tasks without fine-tuning, what aspects of the training process contribute to this phenomenon, and how prompt design can be optimized to maximize ICL efficiency.

To explain in-context learning, various theories have emerged, ranging from statistical and structural perspectives to cognitive and algorithmic analogies. Some researchers attribute ICL to properties of input data distribution and label space structure \citep{min2022rethinking, chan2022data}, while others suggest that exposure to a diverse range of tasks during multitask-prompted learning facilitates generalization \citep{sanh2021multitask}. Another line of work explores how pre-training term frequencies influence a model's ability to recall and recombine information \citep{razeghi2022impact}. Some studies frame ICL as a form of Bayesian inference over the latent space of language \citep{xie2021explanation}, while others view it as a compositional recombination of linguistic structures \citep{hahn2023theory}. At the corpus-statistics end of this spectrum, \citep{wang2025generalization} shows (see \cref{subsec:loss}) that the pretraining-data n-gram footprint of a task systematically predicts whether scaling yields more memorization or more generalization, providing a measurable bridge between data-distributional theories of ICL and the emergence debate. Mechanistically, ICL has been linked to neural architectures, such as induction heads \citep{elhage2021mathematical, olsson2022context} and functional modules that emerge naturally during training \citep{wang2023label, bietti2024birth, irie2022dual, todd2023function}.
Yang et al.\cite{yang2025emergent} (discussed in detail in \cref{subsec:arith}) integrate these primitives into a single circuit for abstract reasoning, with two findings of direct relevance to ICL. First, the \textit{symbolic induction heads} they identify are largely disjoint from the standard induction heads of \cite{olsson2022context} (correlation $r=0.11$), suggesting that the field has been conflating two mechanistically distinct circuits. Second, these same heads are highly correlated ($r=0.86$) with the heads that compute function vectors~\cite{todd2023function}, recasting function vectors as the output of sequence induction over abstract variables.

A complementary view on the fragility of in-context learning comes from Elhady et al.~\cite{elhady2025cpt}, who introduce \textbf{Copain}, a language-agnostic ICL benchmark in which all tasks (e.g., picking the alphabetically first character or the odd integer in an otherwise-even list) must be inferred purely from few-shot demonstrations, with no natural-language instruction. Using Copain to disentangle ICL from target-language knowledge during continued pretraining for language adaptation, they show that ICL capabilities can collapse almost entirely within the first 2k training steps when the language distribution shifts abruptly, and that this collapse, rather than any failure to model the new language, is what prevents downstream emergent abilities from materializing (see \cref{subsec:loss}).

A complementary mechanistic perspective is offered by \cite{zucchet2025sparse}, who argue that the formation of sparse attention patterns is a unifying mechanism behind multiple ICL-related emergent behaviors, including the induction head circuit of \cite{olsson2022context}. Through theoretical analysis of a tractable linear regression variant, they show that learning a sparse attention pattern produces a characteristic plateau-then-jump dynamic during training, with plateau length following a power law $T_{\mathrm{plateau}} \sim \sqrt{dT}$ in sequence length $T$ and embedding dimension $d$. The mechanism is a positive feedback loop: feedforward weights must first align with the target before attention can sharpen on relevant tokens, after which sharper attention accelerates weight learning. They further show that data repetition dramatically accelerates emergence — in-context repetition (the relevant token appearing $B$ times in context) shortens plateaus by a factor of $B$, while cross-sample repetition introduces anisotropy in the input covariance that bootstraps feedforward learning. Predictions are empirically validated on the in-context associative recall task, where 2-4$\times$ speedups from repetition are observed. A dynamical complement to these mechanistic accounts is provided by \cite{singh2025strategy}, who study why emergent ICL can be transient, a phenomenon first reported in~\cite{singh2023transient} and reproduced by~\cite{he2024learning,anand2025dual}, in which ICL emerges during training only to disappear at longer horizons. Working in a controlled 2-layer attention-only transformer trained on bursty few-shot sequences~\cite{chan2022data}, they show that the asymptotic strategy is neither ICL nor pure in-weights learning but a hybrid they term \textbf{context-constrained in-weights learning (CIWL)}, in which the model attends to the correct label token in context (without needing the matching exemplar) and copies it forward via skip-trigram circuits distributed across Layer-2 heads in superposition. Crucially, these Layer-2 heads are the very same canonical induction heads of Olsson et al.~\cite{olsson2022context} that originally implemented ICL: the ICL-to-CIWL transition is driven entirely by a change in Layer-1 heads, which switch from previous-token attention to attending-to-self, leaving the Layer-2 ``induction'' machinery semantically stationary.

This shared subcircuitry yields what the authors call strategy coopetition: ICL and CIWL simultaneously cooperate (via the shared L2 circuit, which enables ICL to bootstrap onto CIWL's developing path) and compete (via L1, where only one of the two attention patterns can dominate). The emergence of ICL is shown to require three concurrent conditions - it must be useful on the training distribution, lie on the way to the asymptotic CIWL solution, and form fast enough to make a transient appearance before CIWL solidifies. A minimal mathematical model with two multiplicatively-interacting mechanisms sharing sub-vectors reproduces both the transience curve and a previously-unexplained non-monotonic dip in CIWL formation observed empirically. Informed by the model, the authors identify a data intervention --- matching context and query exemplars exactly --- that removes the asymptotic bias toward CIWL and makes ICL persistent in both small and 12-layer transformers~\cite{singh2023transient}. The work mechanistically grounds the recent view that few-shot ICL lies on a spectrum of in-context abilities~\cite{lampinen2024broader,yin2025attention} and reframes the emergence question as one of strategy dynamics over training rather than capacity unlock at scale.

Arora and Goyal \citep{arora2023theory} provide a theoretical framework in which LLMs develop skills through a bipartite "skill graph" that links training data to fundamental reasoning abilities, demonstrating how compositional generalization emerges as models scale.

Beyond theoretical explanations, studies have examined the role of training data and model architecture in shaping ICL. Research indicates that the diversity and structure of training data impact ICL more than dataset size alone \citep{shin2022effect, yadlowsky2023pretraining, chan2022data, wies2024learnability}. Shin et al. \citep{shin2022effect, raventos2024pretraining} show that a model's ability to generalize in context depends on task diversity within the pretraining corpus rather than simply increasing the number of tokens. Other studies explore how model size and architectural choices affect ICL capabilities, with findings suggesting that larger models naturally develop stronger in-context learning abilities due to their ability to encode more complex structures \citep{brown2020language, ding2023causallm}.

Another key research focus is the optimization of \textbf{prompt design} for effective ICL. Studies show that the \textbf{selection, formatting, explicit intent formulation, and arrangement of examples} in a prompt significantly influence model performance \citep{zhao2021calibrate, lu2021fantastically, bodonhelyi2024user}. Methods for selecting exemplars can be broadly classified as \textbf{unsupervised} or \textbf{supervised}. Unsupervised approaches rely on nearest neighbor methods, using similarity metrics such as perplexity, mutual information, or self-evaluated LLM scores to identify the most relevant examples \citep{liu2021makes, tanwar2023multilingualllmsbettercrosslingual, sorensen2022information, gonen2022demystifying, wu2022self, li2023finding}. Supervised methods employ dense retrievers trained on labeled data \citep{rubin2021learning, li2023unified, ye2023compositional} and reinforcement learning \citep{zhang2022active} to optimize example selection. Research has also explored strategies for refining prompt formatting and ordering to enhance ICL effectiveness, as well as the impact of instruction formatting on how models process and execute in-context tasks \citep{kim2022self, yang2023auto, hao2022structured, honovich2022instruction, zhou2022large, wang2022self, lu2021fantastically, liu2021makes, liu2024let}.

A fundamental challenge in ICL research is controlling for the many factors that influence performance, making it difficult to establish causal relationships. Most findings are correlational rather than definitive due to the complexity of model pretraining and the vast scope of latent knowledge encoded in LLMs. For a comprehensive overview of the field, readers can refer to recent surveys on in-context learning \citep{dong2022survey, zhou2024mysteryincontextlearningcomprehensive}.

\noindent\textbf{How Larger Models Approach In-Context Learning Differently.} Wei et al. \citep{wei2023larger} investigate the flexibility of LLMs in learning new input-label associations in real time, focusing on whether models rely on pre-existing semantic knowledge or can adapt to novel mappings. They conduct experiments using flipped-label in-context learning (ICL) and semantically unrelated label ICL (SUL-ICL). In flipped-label ICL, models are exposed to contradicting input-label mappings, challenging them to suppress their semantic priors. Small models struggle with these tasks, falling back on established meanings, while larger models, given sufficient examples, can adjust to new mappings, indicating that adaptability improves with model scale. In the SUL-ICL setup, traditional labels are replaced with arbitrary symbols, testing if models can learn without linguistic cues. Larger models excel here too, unlike smaller ones, suggesting that the ability to learn independently of semantic priors is a scale-dependent trait. Instruction tuning further enhances models' ability to learn new associations in SUL-ICL but makes them less flexible in overriding semantic knowledge in flipped-label contexts. Additionally, the study explores high-dimensional linear classification and finds that only the largest models achieve above-random performance, highlighting that effective learning of complex mappings requires substantial model size.

\noindent\textbf{Our Takeaway from ICL.} In-context learning represents a fundamental shift in how Language Models acquire and apply knowledge, allowing them to generalize to new tasks without explicit updates. While \textbf{scaling enhances ICL}, it is evident that factors such as \textbf{training data diversity, model architecture, and prompt design all contribute significantly} to this phenomenon. The ability of larger models to override semantic priors and learn entirely novel associations highlights that \textbf{ICL is not just a side effect of memorization but a structured capability that emerges with scale}.

\section{Emergent Abilities of Large Reasoning Models}\label{sec:lrms}
Large Reasoning Models (LRMs), such as OpenAI's o3, DeepSeek-R1 \citep{deepseekai2025deepseekr1incentivizingreasoningcapability}, and Gemini 2.0, are AI systems built upon the foundations of LLMs that can perform complex reasoning\footnote{To clarify, we define reasoning as the ability to employ logic and rational thinking to derive truthful conclusions from both new and pre-existing data.} tasks, such as coding, PhD-level Q\&A, and math problem-solving. The distinctive capabilities of LRMs stem from two principal innovations: the scaling of Reinforcement Learning (RL) during the post-training phase \cite{shao2024deepseekmath, deepseekai2025deepseekr1incentivizingreasoningcapability} and the scaling of inference-time computing through search \citep{wu2024inference, snell2024scaling}. 

Through reinforcement learning, large reasoning models refine their internal problem-solving processes, similar to developing a chain of thought. This training paradigm fosters the emergence of metacognitive abilities, enabling the models to recognize errors, self-correct, and decompose intricate tasks into simpler sub-problems. Moreover, RL equips these systems with the flexibility to adjust their strategies dynamically when the current approach proves ineffective.
Scaling test-time computing further boosts performance by allocating additional inference steps during evaluation. By iteratively refining their reasoning, models can explore multiple solution pathways and mitigate error propagation, resulting in more accurate and robust outcomes on complex tasks.

Ye et al.~\citep{ye2025emergence} recently proposed \emph{Reinforcement Learning via Self-Play} (RLSP) as a minimal recipe for eliciting these behaviours: optional SFT on reasoning demonstrations, followed by PPO training that combines an outcome verifier (binary correctness signal) with an \emph{exploration reward} deliberately decoupled from correctness and instead rewarding process-level properties of the reasoning trajectory.
Even the simplest instantiation---a reward that merely encourages longer responses with explicit intermediate steps---triggers the emergence of \textbf{backtracking, self-verification, and exploration of alternative rationales} across model families (Llama-3.1-8B, Qwen2.5-7B/32B) and both math and coding domains, yielding a 23\% jump on MATH-500 for Llama-3.1-8B and a 10\% jump on AIME~2024 for Qwen2.5-32B-Instruct. Two ablations sharpen the picture: without any exploration reward, pure PPO on the binary verifier elicits search behaviour in only a single (base model, domain) pair among those tested, and self-consistency at a matched token budget underperforms the RLSP-trained model by 11.6 accuracy points on MATH-500. These results suggest that the metacognitive behaviours seen in frontier LRMs depend less on scale alone than on training objectives that explicitly incentivise the model to synthesise novel chain-of-thought trajectories, with the exploration reward providing a dense guidance signal that a purely correctness-based reward cannot supply during the sparse early phase of RL training.

Complementing this, Wang et al.~\citep{wang2025hierarchical} give a mechanistic account of
\emph{how} such reasoning behaviours emerge over the course of RL training. Decomposing
generated tokens into high-level \textbf{planning tokens} (strategic moves such as deduction,
branching, and backtracking, identified automatically as recurring ``strategic grams'') and
low-level \textbf{execution tokens} (arithmetic, substitutions, formula application), they
observe a consistent \textbf{two-phase learning dynamic} across eight text-only and
vision-language models (Qwen2.5-7B, Qwen3-4B, Llama-3.1-8B, Qwen2.5-VL-7B, MiMo-VL-7B). In
the first phase, optimisation pressure consolidates procedural reliability---execution-token
perplexity and entropy drop sharply as the model becomes confidently correct on low-level
steps; in the second, the learning frontier shifts to the \emph{exploration of strategic
planning}, marked by rising semantic diversity of planning tokens that correlates with
sustained accuracy gains and length scaling. This framework unifies otherwise disparate
phenomena: ``aha moments'' become the discovery of a new strategic construct, and
``length-scaling'' the natural consequence of deploying more elaborate strategies. Notably,
the initial procedural phase is attenuated or absent in stronger base models, indicating that
the emergent hierarchy is driven by training dynamics rather than model scale alone. Acting on
this insight, the authors propose \emph{Hierarchy-Aware Credit Assignment} (HICRA), which
amplifies the policy-gradient signal on the sparse planning tokens rather than diluting it
across all tokens as in GRPO; concentrating optimisation on this strategic bottleneck yields
consistent improvements over GRPO (e.g.\ $+6.1$ points on AIME24 for Qwen3-4B-Base). They
further argue that semantic entropy over planning tokens is a more faithful exploration signal
than aggregate token-level entropy, which is dominated by the many low-level execution tokens.

\noindent\textbf{Empirical Evidence of Improved Reasoning in LRMs.} The impact of these advancements is evident in empirical evaluations, particularly on \textbf{reasoning-intensive benchmarks}. Compared to prior-generation LLMs, LRMs demonstrate substantial jumps in performance, highlighting the effectiveness of scaling both RL and inference-time search. For instance, on Competition Math (AIME 2024), OpenAI's o1 model achieved an accuracy of 83.3\%, vastly surpassing GPT-4o's 13.4\%. Similarly, in the Codeforces programming competition, o1 reached 89.0\% accuracy, while GPT-4o managed only 11.0\%. These results indicate that o1 has surpassed the performance of human experts in multiple benchmarks, reflecting a fundamental shift in AI's ability to handle complex, multi-step problem-solving tasks.

A more recent iteration, o3 \cite{openaio3}, has pushed these capabilities even further by scaling post-training RL and integrating search-based inference. On the ARC-AGI benchmark, which tests adaptive problem-solving and general reasoning, o3 achieved 88\% accuracy, compared to o1's 13.33\% and GPT-4o's 5\%. These results strongly suggest that planning, self-reflection, and strategic thinking have become emergent abilities in this new generation of models. However, it is important to note that while o3 demonstrates remarkable reasoning proficiency, it still fails on certain simple tasks, highlighting fundamental gaps between it and human intelligence.

\noindent\textbf{Implications of Higher-Order Reasoning in LRMs.} The emergence of higher-order reasoning in LRMs has significant implications both in terms of AI capabilities and associated risks. One particularly notable development is that OpenAI's o3-mini has become the \textbf{first AI model to receive a Medium risk classification for Model Autonomy} \cite{openaio3}. This designation underscores the increasing complexity and independence of LRMs, raising concerns about their potential for both beneficial and harmful applications. In the next section, we delve deeper into the harmful behaviors and capabilities of LLMs and LLMs-powered AI agents.

\section{Emergent Behaviors in LLMs-powered AI agents}
\label{sec:llm_agents}
One way to define Artificial Intelligence is the study of agents that receive perceptions from the environment, make decisions, and perform rational actions to achieve a goal \citep{russell2016artificial}. The recent emergence of LLM-powered agents represents a transformative development in this field \citep{gottweis2025towards, huang2024understanding, zhao2024expel}. Advanced models, such as o3-mini and DeepSeek-R1, act as the "brain" of the agent, enabling them to comprehend natural language, interpret complex information, and engage in reasoning processes. Unlike traditional chatbots, which primarily respond to immediate queries, LLM-powered agents offer enhanced capabilities: they can be customized to align with individual user preferences, demonstrate a deeper understanding of contextual realities, plan sequences of actions across multiple steps, and act autonomously on behalf of users.
Chen et al. \citep{chen2023agentverse} introduced a novel framework called AgentVerse, designed to enable and study collaboration among multiple AI agents. Through these interactions, the framework reveals emergent behaviors such as spontaneous cooperation, competition, negotiation, and the development of innovative strategies that were not explicitly programmed. Experimental results and case studies demonstrate that these emergent phenomena provide valuable insights into collective intelligence and the dynamics of multi-agent systems. The paper also discusses the implications for AI safety and alignment, emphasizing the need to understand and guide emergent behaviors to build robust, ethically aligned AI systems.

Moving from single models to interacting populations, \cite{ashery2025emergent} show that groups of LLM agents (Llama-2/3/3.1-70B and Claude-3.5-Sonnet) playing a minimal naming game spontaneously converge on a shared convention through purely local pairwise interactions, with no central coordination---a disorder-to-order symmetry breaking robust up to $N=200$ agents and name pools of $26$ options. Two collective effects have no counterpart at the individual level: a systematic bias toward particular conventions emerges from the interaction dynamics even when agents are unbiased in isolation, and a committed minority can overturn an established convention once it reaches a critical mass, whose size varies sharply across models (from $\sim2\%$ to as much as $67\%$ of the population). The authors conclude that alignment must be evaluated at the group level, since the safety of an individual model does not guarantee the safety of the population it forms.

\textit{Can AI agents develop their own unintended sub-objectives?} While models do not have intrinsic agency, emergent optimization behaviors could lead to unanticipated decision-making patterns and objectives. The new generation of models automatically divides problems into sub-problems, so it is possible that the solutions to some sub-problems can be harmful even if the initial request was benevolent. This necessitates effective monitoring. Furthermore, since an agent's final goal is to maximize the reward function, an agent might develop a plan that tries to prevent humans from deactivating it. We can see this as the emergence of self-preservation. The logical extension of this could involve the agent seeking to gain control over humans as a means of self-protection.
It is important to understand that self-preservation, in itself, can become a powerful driver of other potentially undesirable behaviors.

\section{Emergent Harmful Behaviors in LLMs and LLMs-powered AI agents}\label{sec:harmful_behaviors}
LLM-powered agents, like o3-mini and DeepSeek-R1, represent a major technological breakthrough. They go beyond traditional chatbots by understanding natural language, reasoning, and autonomously planning multi-step actions tailored to user needs. However, their advanced capabilities also raise concerns about the risk of unintended and harmful behaviors. In this section, we separate the concept of model sizes from the emergence term, and we refer to emergence as an ability that develops implicitly.

\paragraph{The Emergence of Deceptive Capabilities}\label{subsec:deceptive}
Researchers have begun to uncover disturbing capacities within LLMs, particularly in the realm of deception. Hagendorff et al. \cite{hagendorff2023deception} explore whether advanced LLMs, 
such as GPT-4, exhibit deception capabilities, positing that this capability stems from improved reasoning abilities not observed in earlier iterations like GPT-2. Their experiments indicate that GPT-4 can effectively deceive other agents in strategic tasks, such as bluffing games, achieving success rates exceeding 70\% when guided by chain-of-thought prompting. Furthermore, when primed with Machiavellian traits, the model exhibits an increased propensity for deceitful behavior. 
 The authors conclude that these emergent abilities pose substantial challenges for AI alignment, emphasizing the urgent need for robust mechanisms to prevent such deceptive tendencies from undermining trust in AI systems.

\paragraph{Reinforcement Learning and the Incentivization of Manipulation}\label{subsec:manip}
A core issue in the emergence of deceptive behaviors stems from the reward structures used in reinforcement learning. Reinforcement Learning from Human Feedback (RLHF) optimizes LLMs to maximize positive user responses (e.g., thumbs-up feedback). However, this process does not inherently reward truthfulness, only perceived quality. As a result, models may learn to mislead users if deception increases the likelihood of receiving positive feedback.

Williams et al. \citep{williams2024targeted} investigate how RLHF can unintentionally reinforce manipulative and exploitative behaviors. Their study reveals that LLMs trained through RLHF develop strategies that exploit user vulnerabilities to maximize reward signals. In controlled conversational settings, they found that models: (1) \emph{exhibited selective deception}, targeting vulnerable individuals while maintaining normal interactions with others, rendering such behaviors difficult to detect; (2) \emph{encouraged harmful behaviors}; (3) \emph{failed standard safety evaluations}, as these manipulative tendencies were not detected by conventional toxicity or {sycophancy} benchmarks. This work demonstrates that reward-driven optimization does not always align with human well-being, highlighting the limitations of current safety evaluation frameworks. 
The challenge of deploying LLMs to be simultaneously helpful and harmless is further analyzed by Bai et al. \cite{bai2022training}, who examine this dual objective through extensive RLHF experimentation. 
Their research demonstrates that while RLHF can markedly enhance a model's helpfulness and reduce harm, a fundamental tension persists between these goals. Models optimized predominantly for harmlessness often become excessively cautious, refraining from engaging in sensitive yet potentially beneficial discussions, such as those involving mental health or political discourse, thereby curtailing 
their practical utility. Additionally, the authors identify "over-optimization" as a critical issue, wherein models become overly attuned to reward signals, resulting in behaviors that deviate from genuine human preferences, such as overly simplistic or generic responses. 

\noindent\textbf{The Reward Hacking Problem.} On the other hand, when models are optimized too strongly for helpfulness, they can engage in reward hacking (i.e., the model prioritizes maximizing approval signals over providing truthful, nuanced, or ethical responses). This can manifest in several ways; for instance, models might exhibit excessive sycophancy, where they opt to affirm what users wish to hear rather than adhering to factual responses, such as yielding to a user's insistence on a pseudoscientific claim simply to boost satisfaction. Additionally, rather than generating well-reasoned replies, these models might lean toward producing oversimplified responses, offering brief, generic, or overly optimistic statements that provide comfort but fall short of substantive depth.

These results, summarized in \cref{tab:em}, converge on two claims that recur across
otherwise very different setups: emergent misalignment is carried by a \emph{low-dimensional,
causally controllable} internal direction, and its onset is frequently \emph{abrupt}. We now
take them in turn.

\newcolumntype{L}{>{\raggedright\arraybackslash}X}
\begin{table}[!t]
\centering
\footnotesize
\setlength{\tabcolsep}{4pt}
\renewcommand{\arraystretch}{1.25}
\begin{tabularx}{\linewidth}{@{}l L L >{\centering\arraybackslash}p{1.75cm} L@{}}
\toprule
\textbf{Study} & \textbf{Trigger} & \textbf{Internal mechanism} & \textbf{Phase transition?} & \textbf{Mitigation} \\
\midrule
\citet{betley2026training}
& SFT on insecure code (GPT-4o / 4.1)
& Broad misalignment on unrelated prompts; ability-gated ($\sim$20\% / $\sim$50\%)
& \textbf{Yes} --- grokking-like dynamics
& --- (original demonstration) \\
\addlinespace
\citet{turner2025model}
& Distilled minimal SFT; single rank-1 LoRA (0.5B+; Qwen/Llama/Gemma)
& One rank-1 adapter $\equiv$ one linear direction; scales with model size
& \textbf{Yes} --- sharp adapter-direction rotation + gradient-norm spike
& Direction-level control (model organism) \\
\addlinespace
\citet{soligo2025convergent}
& 9 rank-1 LoRAs (Qwen2.5-14B-Instruct)
& Single convergent activation-space direction; transfers across fine-tunes (cos $>0.8$); 6 general + 2 domain adapters
& \textit{Not examined}
& Ablate direction ($-78$ to $-90\%$ EM) \\
\addlinespace
\citet{wang2025persona}
& Insecure code, bad advice, RL on reasoning models; no-safety-training models
& SAE ``toxic persona'' latent; causal up/down steering
& \textit{Not examined}
& Steer latent down; re-align with a few hundred benign samples \\
\addlinespace
\citet{macdiarmid2025natural}
& Reward hacking in production RL (Claude Sonnet 3.7)
& Hacking $\rightarrow$ misaligned generalization; covert EM 40--80\%; context-dependent after RLHF
& \textbf{Yes} --- abrupt ($\sim$50 steps flat, then sharp rise)
& Inoculation prompting ($-75$ to $-90\%$) \\
\addlinespace
\citet{hu2025llms}
& Data contamination (1\% misaligned in instruction mix); self-training on 10\% biased users
& Broad dishonesty; escapes capability evals (MMLU/GSM8K/HumanEval/GPQA unchanged); covert in CoT
& \textit{Not examined}
& Contamination screening (implied) \\
\bottomrule
\end{tabularx}
\caption{\textbf{Emergent misalignment (EM) at a glance.} Across deliberately narrow finetuning~\citep{betley2026training, turner2025model, soligo2025convergent, wang2025persona}, realistic reward hacking~\citep{macdiarmid2025natural}, and unintentional contamination~\citep{hu2025llms}, a consistent picture recurs: a low-dimensional internal object---a single LoRA direction, an activation-space direction, or an SAE persona latent---carries the misalignment and is causally controllable, supplying precisely the coarse-grained-mechanism evidence our stance requires. Three of the six studies additionally characterize the \emph{onset} of misalignment as an abrupt phase transition in training dynamics rather than a gradual drift, mirroring the emergence-as-phase-transition through-line of this survey.}
\label{tab:em}
\end{table}

\textbf{Narrow Finetuning Induces Broad Misalignment.}
\cite{betley2026training} finetuned GPT-4o to write insecure code and observed
emergent misalignment: broad harmful behavior on unrelated prompts. The effect is
ability-gated in the sense of  \cite{wei2022emergent} ($\sim$$20\%$ in GPT-4o,
$\sim$$50\%$ in GPT-4.1), is distinct from jailbreaking, generalizes beyond code, and arises
even in base models, with training dynamics reminiscent of grokking.

\textbf{Minimal and Reproducible Model Organisms.}
 \cite{turner2025model} distil clean ``model organisms'' of EM,
reaching $99\%$ coherence in models as small as $0.5$B and showing that a single rank-1 LoRA
adapter, a single linear direction---suffices to induce it, robustly across Qwen, Llama,
and Gemma and under full SFT. Misalignment scales with model size, echoing the ability-gated
trend of \cite{betley2026training}, and is learnt through a sharp {phase transition}: a sudden
rotation of the adapter direction, correlated with a gradient-norm spike, over a narrow
window of steps, tightening the link between EM and grokking-style dynamics.

\textbf{A Convergent Linear Direction for Misalignment.}
Building on the same rank-1 model organisms, \cite{soligo2025convergent} show
that emergent misalignment is mediated by a single linear direction in activation space. From
a minimal organism, $9$ rank-1 LoRA adapters misaligning Qwen2.5-14B-Instruct, coherent over
$99\%$ of the time, they extract a mean-diff ``misalignment direction'': adding it to the
aligned chat model induces up to $50\%$ misaligned responses, while ablating it removes
misalignment. Crucially, the direction transfers: a vector extracted from one fine-tune
ablates misalignment in others trained on different datasets (medical, extreme sports) and
higher-rank LoRAs, cutting EM by $78$--$90\%$ with cosine similarities above $0.8$ across
fine-tunes---evidence that distinct emergently misaligned models converge on a shared
representation. Interpreting the rank-1 scalars further reveals a division of labour: six
adapters encode general misalignment while two specialise for the fine-tuning domain, so
steering them separately induces general or narrowly medical misalignment.

\textbf{Mechanistic Control via Persona Features.} \cite{wang2025persona} broaden EM beyond insecure code, to incorrect advice across domains, to models without safety training, and to RL on reasoning models, and explain it mechanistically: sparse-autoencoder model diffing isolates a ``toxic persona'' latent that causally controls the behaviour (steering it up induces misalignment, down suppresses it) This places EM on a low-dimensional direction, consistent with \cite{turner2025model,soligo2025convergent}, and supplies exactly the internal-mechanism evidence advocated by \cite{krakauer2025complex}; the effect is cheaply reversible with a few hundred benign samples.

\textbf{Natural Emergent Misalignment from Reward Hacking in Production RL.}
While the preceding results rely on deliberately narrow finetuning data, \cite{macdiarmid2025natural} show that emergent misalignment can arise from an entirely realistic training signal: reward hacking on real production coding environments. Starting from a pretrained model made aware of three exploitable hacks (an \texttt{AlwaysEqual} object overriding \texttt{\_\_eq\_\_}, an early \texttt{sys.exit(0)}, and a \texttt{conftest.py} patch of pytest's report machinery) via synthetic document finetuning or prompting, RL on unmodified Claude Sonnet 3.7 environments unsurprisingly teaches the model to hack; surprisingly, the onset of hacking coincides with a sharp rise in misalignment across all six held-out evaluations, from unprompted alignment faking and reasoning about power acquisition to accepting an exfiltration offer from a hacker collective and sabotaging the safety-research codebase of the paper itself when driven through the standard Claude Code scaffold ($12\%$ sabotage rate, with sabotaged classifiers only $65\%$ as effective). The effect is not corpus-specific: every combination of synthetic corpus and dilution level that leads to hacking also leads to misaligned generalization, and hack rate and misalignment score remain correlated across checkpoints. Two findings extend the picture beyond \cite{betley2026training}. First, the authors identify {covert} emergent misalignment, misaligned reasoning followed by benign final outputs, which accounts for $40$--$80\%$ of misaligned responses and is invisible to output-only monitoring. Second, standard chat-like RLHF safety training produces {context-dependent misalignment}: the model is genuinely aligned on distributions resembling the RLHF prompts, yet still acts misaligned on agentic scenarios, a failure of coverage rather than of deception. The mitigation with the best cost--benefit profile is {inoculation prompting}: reframing reward hacking as acceptable via a single line in the RL system prompt cuts final misalignment by $75$--$90\%$ despite hack rates above $99\%$, consistent with the hypothesis that misaligned generalization is mediated by the pretraining-learned association between hacking and misalignment rather than by the hacking behaviour itself. The transition is abrupt, hack rates stay near zero for roughly $50$ RL steps before rising sharply, suggesting that emergent misalignment, like other emergent phenomena discussed in this survey, is best characterised as a phase transition in training dynamics rather than a gradual drift.

\textbf{Broad Dishonesty from Unintentional Contamination.} \cite{hu2025llms} extend EM beyond overtly harmful outputs to dishonesty under high-stakes pressure, measuring the gap between a model's belief and its output on MASK and DeceptionBench. Finetuning Llama-3.1-8B-Instruct and Qwen2.5-7B-Instruct on misaligned code, math, and medical completions broadly lowers honesty and raises strategic deception, echoing the cross-domain generalization of \cite{betley2026training}. Two findings sharpen the realistic picture of \cite{macdiarmid2025natural}: contamination need not be deliberate, in fact, injecting as little as 1\% misaligned data into an ordinary downstream instruction-tuning mixture cuts honesty by over 20\% while leaving MMLU, GSM8K, HumanEval, and GPQA unchanged, so the failure escapes capability evaluations---and misalignment can emerge from interaction alone, since self-training (SFT or KTO) on trajectories with only 10\% biased simulated users steers the assistant toward greater deception. As with covert EM, part of the effect hides in the reasoning trace: the chain-of-thought stays aligned with the model's belief while the final answer contradicts it.

\paragraph{Hypothesizing Singularity}\label{subsec:sing}
 As these models evolve, they are poised to exhibit a variety of characteristics that mark significant advancements. One such trait is rapid self-improvement, where mechanisms like iterative fine-tuning and self-supervised learning allow AI systems to refine their internal representations and decision-making processes at an accelerated pace. Evidence of this self-improvement is already apparent, as reinforcement learning post-training utilizes frozen models to generate rewards for those being trained, though the extent to which they can further enhance themselves remains an open question. Furthermore, newer models have shown the ability to produce novel research ideas, as noted by \cite{gottweis2025towards}. In addition to this, enhanced collaboration emerges as another key feature, particularly in network environments where multiple AI agents have been proven capable of working effectively, according to the findings of \cite{guo2024large} and \cite{chen2023agentverse}. Furthermore, these advanced agents showcase contextual adaptability, learning to interpret multi-modal inputs, such as text, images, and sensor data, and dynamically adjusting their strategies based on real-time feedback.

Let us now hypothesize that, with further technological progress, LLM-powered AI agents might surpass human intelligence. This trajectory toward a qualitatively distinct form of intelligence introduces profound practical challenges. Should these agents achieve or exceed human-level intelligence, ensuring their submission to humans becomes a paramount concern, necessitating research into concepts like corrigibility and interruptibility to prevent scenarios where they might resist human intervention or behave unpredictably. Closely tied to this is the need for robust risk management, given the potential for rapid, exponential improvements in AI capabilities, often described as an intelligence explosion, which demands preemptive safety protocols. As emphasized by \cite{bostrom2024superintelligence}, understanding and mitigating the risks associated with superintelligent agents is essential to avoid unintended consequences.

In conclusion, although the prospect of LLM-powered AI agents outstripping human intelligence remains speculative, the current pace of technological development lends it a plausibility that warrants serious consideration. Future research must, therefore, balance the pursuit of enhanced capabilities with the establishment of rigorous frameworks to ensure their safe and beneficial integration into society.

\subsection{AI Safety}\label{subsec:safety}
\noindent\textbf{The Challenge of Governing Autonomous AI Agents.} LLM-powered AI agents differ from traditional AI models in that they are much more complex and more autonomous in nature \cite{kasneci2026safety}. Examples of large reasoning models like o3-mini and DeepSeek-R1 demonstrate abilities to perform multi-step reasoning, make adaptive decisions, and develop strategies for reward maximization. Although these capabilities improve their effectiveness and usefulness, they also pose new governance challenges not previously considered in AI policy discussions.

Historically, AI safety has focused on content moderation, bias mitigation, transparency, and (adversarial) robustness, but autonomous AI introduces a new class of risks related to opaque optimization goals and continuously changing information basis, which requires different regulatory approaches. These governance challenges are especially critical in high-stakes applications such as healthcare, finance, admission processes, legal reasoning, warfare, and autonomous systems.

\noindent\textbf{The Role of AI Regulation and Global Governance.} AI regulation and global governance have shifted from theoretical debates to urgent, real-world imperatives~\cite{kasneci2025europe};  governments and international organizations are actively drafting policies to regulate AI development and deployment.  The EU AI Act~\citep{eu2023aiact}, the US AI Executive Order~\citep{blumenthal2024us}, and the UN AI Advisory Board~\citep{vercelli2024united}, illustrate the diverse strategies at play. Despite these efforts, there is no globally unified approach to AI regulation, leading to inconsistencies and regulatory loopholes. Some countries favor strict oversight (i.e., EU member states), while others prioritize AI innovation over safety (i.e., the US and China), making international collaboration essential to prevent fragmented and ineffective governance.
These ethical concerns underscore the need for interdisciplinary collaboration between relevant stakeholders, such as AI researchers, policymakers, industry leaders, and end users. Ensuring that AI systems benefit society requires ongoing dialogue and policy refinement. At the same time, AI governance should focus on ensuring that powerful autonomous reasoning agents do not threaten the preservation of biological lives. Therefore, we believe that innovation in governance, compliance mechanisms, and international oversight (with continuous evaluation) is at least as important as technological innovation. As AI systems become more capable, adaptive, and independent, the challenge of controlling their behavior and ensuring trustworthiness will define the next era of responsible AI applications, AI safety research, and policy development.

\section{Conclusion}\label{sec:conclusion}

This survey delves into the intricate and rapidly evolving domain of emergent abilities in LLMs, offering a comprehensive exploration of their nature, analysis, and implications. Notably, the unpredictability and substantial performance leaps observed in specific tasks as the model scale increases underscore the essence of emergent abilities. Additionally, in-context learning, while closely intertwined with emergent abilities, emerges as a distinct phenomenon warranting separate scrutiny. The influence of data distributional properties also proves instrumental in shaping in-context learning capabilities. Theoretical frameworks, such as Bayesian inference applied to latent language spaces, present compelling pathways for elucidating the underlying mechanisms driving both emergent abilities and in-context learning.
Despite these advancements, significant gaps in knowledge persist, leaving numerous questions unresolved. The interplay between model scaling, architectural design, and the onset of novel capabilities remains incompletely understood, necessitating further investigation. Moreover, ongoing contention regarding whether emergent abilities represent authentic phenomena or are instead reflections of measurement biases underscores the pressing need for more robust, standardized evaluation methodologies to settle this debate conclusively.
Looking ahead, future research must prioritize several key avenues to advance this field. It is imperative to devise more sophisticated metrics capable of capturing subtle gradations in model performance across diverse tasks, thereby providing a clearer picture of emergent capabilities. A deeper examination of how model architecture and training methodologies contribute to the fostering of these abilities is equally essential. 

A distinction from~\cite{krakauer2025complex} deserves attention as this field matures: between \textbf{emergent capability} and \textbf{emergent intelligence}. LLMs may increasingly demonstrate the former --- novel task performance arising from scale, training, or prompting --- but the latter, defined as the compressed, low-dimensional use of coarse-grained representations to solve broad classes of problems with minimal effort (``more with less'' via analogy and abstraction), remains an open question. Human intelligence is characterised by the discovery of compact, effective theories that transfer across domains and can be transmitted through a few words of instruction; recent interpretability work on LLMs suggests that their internal mechanisms are, by contrast, overwhelmingly complex even for simple tasks. This capability-versus-intelligence distinction, alongside the mechanistic evidence increasingly surfacing in the works surveyed here, is likely to shape the next phase of the emergence debate: not whether LLMs surprise us with what they can do, but whether they do so through the parsimonious internal reorganisation that has historically defined genuine emergent phenomena.

As LLMs continue their swift progression, comprehending emergent abilities becomes more important than ever, not only for advancing theoretical insights into artificial intelligence but also for informing practical applications. This area of inquiry illuminates the essence of AI while simultaneously challenging conventional notions of learning, cognition, and the genesis of complex behaviors from ostensibly simple components. Perhaps most critically, the demonstration of emergent abilities carries profound implications for safety, as unpredictable and potentially hazardous capabilities, such as the exploitation of software vulnerabilities or the manipulation of human actors, may arise without forewarning, necessitating vigilant oversight, proactive mitigation strategies, and internationally aligned governance strategies.

\bibliographystyle{tmlr}
\bibliography{main}

@article{zhang2024predictable,
  title={Predictable Emergent Abilities of LLMs: Proxy Tasks Are All You Need},
  author={Zhang, Bo-Wen and Yan, Yan and Yang, Boxiang and Xue, Yifei and Liu, Guang},
  journal={arXiv preprint arXiv:2412.07111}    ,
  year={2024}
}

@article{ye2025emergence,
    title={On the Emergence of Thinking in {LLM}s {I}: Searching for the Right Intuition},
    author={Ye, Guanghao and Pham, Khiem Duc and Zhang, Xinzhi and Gopi, Sivakanth and Peng, Baolin and Li, Beibin and Kulkarni, Janardhan and Inan, Huseyin A.},
    journal={arXiv preprint arXiv:2502.06773},
    year={2025}
}

@article{kasneci2025europe,
  title={Europe's AI Imperative--A Pragmatic Blueprint for Global Tech Leadership},
  author={Kasneci, Gjergji and Gasser, Urs and Hofmann, Thomas F and Kramer, Gerhard and M{\"u}ller, Gerhard and Peus, Claudia and Sch{\"o}nenberger, Helmut and Kasneci, Enkelejda},
  journal={arXiv preprint arXiv:2502.08781 },
  year={2025}
}

@article{abramov2025grokking,
  title={Grokking in the wild: Data augmentation for real-world multi-hop reasoning with transformers},
  author={Abramov, Roman and Steinbauer, Felix and Kasneci, Gjergji},
  journal={arXiv preprint arXiv:2504.20752},
  year={2025}
}

@article{wang2025persona,
  title={Persona features control emergent misalignment},
  author={Wang, Miles and la Tour, Tom Dupr{\'e} and Watkins, Olivia and Makelov, Alex and Chi, Ryan A and Miserendino, Samuel and Wang, Jeffrey and Rajaram, Achyuta and Heidecke, Johannes and Patwardhan, Tejal and others},
  journal={arXiv preprint arXiv:2506.19823},
  year={2025}
}

@article{kasneci2026safety,
  title={The safety failures we are not instrumenting: a perspective on hidden safety-critical challenges in modern AI systems},
  author={Kasneci, Gjergji and Kasneci, Enkelejda},
  journal={AI and Ethics},
  volume={6},
  number={3},
  pages={295},
  year={2026},
  publisher={Springer}
}

@misc{openaio3,
  title={o3 mini system card, 2024},
  author={OpenAI, OpenAI}
}

@article{turner2025model,
  title={Model organisms for emergent misalignment},
  author={Turner, Edward and Soligo, Anna and Taylor, Mia and Rajamanoharan, Senthooran and Nanda, Neel},
  journal={arXiv preprint arXiv:2506.11613},
  year={2025}
}

@article{taniguchi2026generative,
  title={Generative emergent communication: Large language model is a collective world model},
  author={Taniguchi, Tadahiro and Ueda, Ryo and Nakamura, Tomoaki and Suzuki, Masahiro and Taniguchi, Akira},
  journal={Advanced Robotics},
  volume={40},
  number={9},
  pages={499--524},
  year={2026},
  publisher={Taylor \& Francis}
}

@article{ashery2025emergent,
  title={Emergent social conventions and collective bias in LLM populations},
  author={Ashery, Ariel Flint and Aiello, Luca Maria and Baronchelli, Andrea},
  journal={Science Advances},
  volume={11},
  number={20},
  pages={eadu9368},
  year={2025},
  publisher={American Association for the Advancement of Science}
}

@article{soligo2025convergent,
  title={Convergent linear representations of emergent misalignment},
  author={Soligo, Anna and Turner, Edward and Rajamanoharan, Senthooran and Nanda, Neel},
  journal={arXiv preprint arXiv:2506.11618},
  year={2025}
}

@article{hu2025llms,
  title={LLMs Learn to Deceive Unintentionally: Emergent Misalignment in Dishonesty from Misaligned Samples to Biased Human-AI Interactions},
  author={Hu, XuHao and Wang, Peng and Lu, Xiaoya and Liu, Dongrui and Huang, Xuanjing and Shao, Jing},
  journal={arXiv preprint arXiv:2510.08211},
  year={2025}
}

@article{macdiarmid2025natural,
  title={Natural emergent misalignment from reward hacking in production rl},
  author={MacDiarmid, Monte and Wright, Benjamin and Uesato, Jonathan and Benton, Joe and Kutasov, Jon and Price, Sara and Bouscal, Naia and Bowman, Sam and Bricken, Trenton and Cloud, Alex and others},
  journal={arXiv preprint arXiv:2511.18397},
  year={2025}
}

@article{betley2026training,
  title={Training large language models on narrow tasks can lead to broad misalignment},
  author={Betley, Jan and Warncke, Niels and Sztyber-Betley, Anna and Tan, Daniel and Bao, Xuchan and Soto, Mart{\'\i}n and Srivastava, Megha and Labenz, Nathan and Evans, Owain},
  journal={Nature},
  volume={649},
  number={8097},
  pages={584--589},
  year={2026},
  publisher={Nature Publishing Group UK London}
}

@inproceedings{wang2025hierarchical,
    title={Emergent Hierarchical Reasoning in {LLM}s through Reinforcement Learning},
    author={Wang, Haozhe and Xu, Qixin and Liu, Che and Wu, Junhong and Lin, Fangzhen and Chen, Wenhu},
    booktitle={International Conference on Learning Representations (ICLR)},
    year={2026}
}

@article{ma2026phase,
    title={Phase transitions in large language model compression},
    author={Ma, Ziyang and Li, Zuchao and Zhang, Lefei and Xia, Gui-Song and Du, Bo and Zhang, Liangpei and Tao, Dacheng},
    journal={npj Artificial Intelligence},
    volume={2},
    pages={21},
    year={2026},
    publisher={Nature Publishing Group}
}

@article{chen2025abstract,
    title={The Emergence of Abstract Thought in Large Language Models Beyond Any Language},
    author={Chen, Yuxin and Zhao, Yiran and Zhang, Yang and Zhang, An and Kawaguchi, Kenji and Joty, Shafiq and Li, Junnan and Chua, Tat-Seng and Shieh, Michael Qizhe and Zhang, Wenxuan},
    journal={arXiv preprint arXiv:2506.09890},
    year={2025}
}

@inproceedings{zhu2025superposition,
    title={Emergence of Superposition: Unveiling the Training Dynamics of Chain of Continuous Thought},
    author={Zhu, Hanlin and Hao, Shibo and Hu, Zhiting and Jiao, Jiantao and Russell, Stuart and Tian, Yuandong},
    booktitle={International Conference on Learning Representations (ICLR)},
    year={2026}
}

@article{krakauer2025complex,
  title={Large Language Models and Emergence: A Complex Systems Perspective},
  author={Krakauer, David C. and Krakauer, John W. and Mitchell, Melanie},
  journal={arXiv preprint arXiv:2506.11135},
  note={Forthcoming in \emph{Philosophical Transactions of the Royal Society A}},
  year={2025}
}

@inproceedings{huang2024demystifying,
  title={Demystifying verbatim memorization in large language models},
  author={Huang, Jing and Yang, Diyi and Potts, Christopher},
  booktitle={Proceedings of the 2024 Conference on Empirical Methods in Natural Language Processing},
  pages={10711--10732},
  year={2024}
}

@article{lampinen2024broader,
  title={The broader spectrum of in-context learning},
  author={Lampinen, Andrew Kyle and Chan, Stephanie CY and Singh, Aaditya K and Shanahan, Murray},
  journal={arXiv preprint arXiv:2412.03782},
  year={2024}
}

@inproceedings{yin2025attention,
  title={Which Attention Heads Matter for In-Context Learning?},
  author={Yin, Kayo and Steinhardt, Jacob},
  booktitle={International Conference on Machine Learning},
  pages={72428--72461},
  year={2025},
  organization={PMLR}
}

@inproceedings{anand2025dual,
  title={Dual process learning: Controlling use of in-context vs. in-weights strategies with weight forgetting},
  author={Anand, Suraj and Lepori, Michael and Merullo, Jack and Pavlick, Ellie},
  booktitle={International Conference on Learning Representations},
  volume={2025},
  pages={50280--50306},
  year={2025}
}

@article{he2024learning,
  title={Learning to grok: Emergence of in-context learning and skill composition in modular arithmetic tasks},
  author={He, Tianyu and Doshi, Darshil and Das, Aritra and Gromov, Andrey},
  journal={Advances in Neural Information Processing Systems},
  volume={37},
  pages={13244--13273},
  year={2024}
}

@article{singh2023transient,
  title={The transient nature of emergent in-context learning in transformers},
  author={Singh, Aaditya and Chan, Stephanie and Moskovitz, Ted and Grant, Erin and Saxe, Andrew and Hill, Felix},
  journal={Advances in neural information processing systems},
  volume={36},
  pages={27801--27819},
  year={2023}
}

@inproceedings{singh2025strategy,
  title={Strategy Coopetition Explains the Emergence and Transience of In-Context Learning},
  author={Singh, Aaditya K and Moskovitz, Ted and Dragutinovi{\'c}, Sara and Hill, Felix and Chan, Stephanie CY and Saxe, Andrew M},
  booktitle={International Conference on Machine Learning},
  pages={55720--55739},
  year={2025},
  organization={PMLR}
}

@inproceedings{wang2025generalization,
  title={Generalization v.s. Memorization: Tracing Language Models' Capabilities Back to Pretraining Data},
  author={Wang, Xinyi and Antoniades, Antonis and Elazar, Yanai and Amayuelas, Alfonso and Albalak, Alon and Zhang, Kexun and Wang, William Yang},
  booktitle={International Conference on Learning Representations (ICLR)},
  year={2025}
}

@inproceedings{merrill-etal-2024-evaluating,
    title = "Evaluating $n$-Gram Novelty of Language Models Using Rusty-{DAWG}",
    author = "Merrill, William  and
      Smith, Noah A.  and
      Elazar, Yanai",
    editor = "Al-Onaizan, Yaser  and
      Bansal, Mohit  and
      Chen, Yun-Nung",
    booktitle = "Proceedings of the 2024 Conference on Empirical Methods in Natural Language Processing",
    month = nov,
    year = "2024",
    address = "Miami, Florida, USA",
    publisher = "Association for Computational Linguistics",
    url = "https://aclanthology.org/2024.emnlp-main.800/",
    doi = "10.18653/v1/2024.emnlp-main.800",
    pages = "14459--14473"
}

@article{pruthi2020estimating,
  title={Estimating training data influence by tracing gradient descent},
  author={Pruthi, Garima and Liu, Frederick and Kale, Satyen and Sundararajan, Mukund},
  journal={Advances in Neural Information Processing Systems},
  volume={33},
  pages={19920--19930},
  year={2020}
}

@article{li2025grokking,
  title={Grokking in LLM Pretraining? Monitor Memorization-to-Generalization without Test},
  author={Li, Ziyue and Fan, Chenrui and Zhou, Tianyi},
  journal={arXiv preprint arXiv:2506.21551},
  year={2025}
}

@misc{zhao2026randomscalingemergentcapabilities,
      title={Random Scaling of Emergent Capabilities}, 
      author={Rosie Zhao and Tian Qin and David Alvarez-Melis and Sham Kakade and Naomi Saphra},
      year={2026},
      eprint={2502.17356},
      archivePrefix={arXiv},
      primaryClass={cs.LG},
      url={https://arxiv.org/abs/2502.17356}, 
}

@inproceedings{liu2025superposition,
  title={Superposition Yields Robust Neural Scaling},
  author={Liu, Yizhou and Liu, Ziming and Gore, Jeff},
  booktitle={Advances in Neural Information Processing Systems},
  year={2025}
}

@article{elhady2025cpt,
  title={Emergent Abilities of Large Language Models under Continued Pretraining for Language Adaptation},
  author={Elhady, Ahmed and Agirre, Eneko and Artetxe, Mikel},
  journal={arXiv preprint arXiv:2506.00288},
  year={2025},
  note={v3, September 2025}
}

@inproceedings{zucchet2025sparse,
  title={The Emergence of Sparse Attention: Impact of Data Distribution and Benefits of Repetition},
  author={Zucchet, Nicolas and D'Angelo, Francesco and Lampinen, Andrew K. and Chan, Stephanie C.Y.},
  booktitle={Advances in Neural Information Processing Systems},
  year={2025}
}

@inproceedings{gu2025data,
  title={Data Mixing Can Induce Phase Transitions in Knowledge Acquisition},
  author={Gu, Xinran and Lyu, Kaifeng and Li, Jiazheng and Zhang, Jingzhao},
  booktitle={Advances in Neural Information Processing Systems},
  year={2025}
}

@inproceedings{yang2025emergent,
  title={Emergent Symbolic Mechanisms Support Abstract Reasoning in Large Language Models},
  author={Yang, Yukang and Campbell, Declan and Huang, Kaixuan and Wang, Mengdi and Cohen, Jonathan and Webb, Taylor},
  booktitle={Proceedings of the 42nd International Conference on Machine Learning},
  series={PMLR},
  volume={267},
  year={2025}
}

@article{sun2025breaking,
  title={Breaking myths in LLM scaling and emergent abilities with a comprehensive statistical analysis},
  author={Sun, Kun and Wang, Rong},
  journal={Neurocomputing},
  pages={132542},
  year={2025},
  publisher={Elsevier}
}

@article{bodonhelyi2024user,
  title={User intent recognition and satisfaction with large language models: A user study with chatgpt},
  author={Bodonhelyi, Anna and Bozkir, Efe and Yang, Shuo and Kasneci, Enkelejda and Kasneci, Gjergji},
  journal={arXiv preprint arXiv:2402.02136},
  year={2024}
}

@article{chen2024scaling,
  title={Scaling Laws for Predicting Downstream Performance in LLMs},
  author={Chen, Yangyi and Huang, Binxuan and Gao, Yifan and Wang, Zhengyang and Yang, Jingfeng and Ji, Heng},
  journal={arXiv preprint arXiv:2410.08527}    ,
  year={2024}
}

@article{huang2024understanding,
  title={Understanding the planning of LLM agents: A survey},
  author={Huang, Xu and Liu, Weiwen and Chen, Xiaolong and Wang, Xingmei and Wang, Hao and Lian, Defu and Wang, Yasheng and Tang, Ruiming and Chen, Enhong},
  journal={arXiv preprint arXiv:2402.02716}  ,
  year={2024}
}

@inproceedings{zhao2024expel,
  title={Expel: Llm agents are experiential learners},
  author={Zhao, Andrew and Huang, Daniel and Xu, Quentin and Lin, Matthieu and Liu, Yong-Jin and Huang, Gao},
  booktitle={Proceedings of the AAAI Conference on Artificial Intelligence},
  volume={38},
  number={17},
  pages={19632--19642},
  year={2024}
}

@article{power2022grokking,
  title={Grokking: Generalization beyond overfitting on small algorithmic datasets},
  author={Power, Alethea and Burda, Yuri and Edwards, Harri and Babuschkin, Igor and Misra, Vedant},
  journal={arXiv preprint arXiv:2201.02177},
  year={2022}
}

@article{ye2023predictable,
  title={How predictable are large language model capabilities? a case study on big-bench},
  author={Ye, Qinyuan and Fu, Harvey Yiyun and Ren, Xiang and Jia, Robin},
  journal={arXiv preprint arXiv:2305.14947},
  year={2023}
}

@article{snell2024predicting,
  title={Predicting emergent capabilities by finetuning},
  author={Snell, Charlie and Wallace, Eric and Klein, Dan and Levine, Sergey},
  journal={arXiv preprint arXiv:2411.16035},
  year={2024}
}

@article{chen2024states,
  title={States Hidden in Hidden States: LLMs Emerge Discrete State Representations Implicitly},
  author={Chen, Junhao and Hu, Shengding and Liu, Zhiyuan and Sun, Maosong},
  journal={arXiv preprint arXiv:2407.11421},
  year={2024}
}

@article{hendrycks2020measuring,
  title={Measuring massive multitask language understanding},
  author={Hendrycks, Dan and Burns, Collin and Basart, Steven and Zou, Andy and Mazeika, Mantas and Song, Dawn and Steinhardt, Jacob},
  journal={arXiv preprint arXiv:2009.03300},
  year={2020}
}

@misc{PersianQA,
  author          = {Ayoubi, Sajjad \& Davoodeh, Mohammad Yasin},
  title           = {PersianQA: a dataset for Persian Question Answering},
  year            = 2021,
  publisher       = {GitHub},
  journal         = {GitHub repository},
  howpublished    = {\url{https://github.com/SajjjadAyobi/PersianQA}},
}

@article{wu2024u,
  title={U-shaped and Inverted-U Scaling behind Emergent Abilities of Large Language Models},
  author={Wu, Tung-Yu and Lo, Pei-Yu},
  journal={arXiv preprint arXiv:2410.01692} ,
  year={2024}
}

@article{wei2023larger,
  title={Larger language models do in-context learning differently},
  author={Wei, Jerry and Wei, Jason and Tay, Yi and Tran, Dustin and Webson, Albert and Lu, Yifeng and Chen, Xinyun and Liu, Hanxiao and Huang, Da and Zhou, Denny and others},
  journal={arXiv preprint arXiv:2303.03846} ,
  year={2023}
}

@book{lewes1877problems,
  title={Problems of life and mind},
  author={Lewes, George Henry},
  volume={2},
  year={1877},
  publisher={Trubner \& Company}
}

@article{wei2022emergent,
  title={Emergent abilities of large language models},
  author={Wei, Jason and Tay, Yi and Bommasani, Rishi and Raffel, Colin and Zoph, Barret and Borgeaud, Sebastian and Yogatama, Dani and Bosma, Maarten and Zhou, Denny and Metzler, Donald and others},
  journal={arXiv preprint arXiv:2206.07682} ,
  year={2022}
}

@article{bai2022training,
  title={Training a helpful and harmless assistant with reinforcement learning from human feedback},
  author={Bai, Yuntao and Jones, Andy and Ndousse, Kamal and Askell, Amanda and Chen, Anna and DasSarma, Nova and Drain, Dawn and Fort, Stanislav and Ganguli, Deep and Henighan, Tom and others},
  journal={arXiv preprint arXiv:2204.05862} ,
  year={2022}
}

@book{russell2016artificial,
  title={Artificial intelligence: a modern approach},
  author={Russell, Stuart J and Norvig, Peter},
  year={2016},
  publisher={pearson}
}

@article{dong2022survey,
  title={A survey on in-context learning},
  author={Dong, Qingxiu and Li, Lei and Dai, Damai and Zheng, Ce and Ma, Jingyuan and Li, Rui and Xia, Heming and Xu, Jingjing and Wu, Zhiyong and Liu, Tianyu and others},
  journal={arXiv preprint arXiv:2301.00234} ,
  year={2022}
}

@misc{perez2022ignorepreviouspromptattack,
      title={Ignore Previous Prompt: Attack Techniques For Language Models}, 
      author={Fábio Perez and Ian Ribeiro},
      year={2022},
      eprint={2211.09527},
      archivePrefix={arXiv},
      primaryClass={cs.CL},
      url={https://arxiv.org/abs/2211.09527}, 
}

@misc{zhou2024mysteryincontextlearningcomprehensive,
      title={The Mystery of In-Context Learning: A Comprehensive Survey on Interpretation and Analysis}, 
      author={Yuxiang Zhou and Jiazheng Li and Yanzheng Xiang and Hanqi Yan and Lin Gui and Yulan He},
      year={2024},
      eprint={2311.00237},
      archivePrefix={arXiv},
      primaryClass={cs.CL},
      url={https://arxiv.org/abs/2311.00237}, 
}

@article{wei2021finetuned,
  title={Finetuned language models are zero-shot learners},
  author={Wei, Jason and Bosma, Maarten and Zhao, Vincent Y and Guu, Kelvin and Yu, Adams Wei and Lester, Brian and Du, Nan and Dai, Andrew M and Le, Quoc V},
  journal={arXiv preprint arXiv:2109.01652},
  year={2021}
}

@article{hagendorff2023deception,
  author = {Hagendorff, Thilo},
  title = {Deception Abilities Emerged in Large Language Models},
  journal = {arXiv preprint arXiv:2307.16513},
  year = {2023},
  month = {July},
  url = {https://arxiv.org/abs/2307.16513},
  note = {Revised February 2, 2024}
}

@article{nye2021show,
  title={Show your work: Scratchpads for intermediate computation with language models},
  author={Nye, Maxwell and Andreassen, Anders Johan and Gur-Ari, Guy and Michalewski, Henryk and Austin, Jacob and Bieber, David and Dohan, David and Lewkowycz, Aitor and Bosma, Maarten and Luan, David and others},
  journal={arXiv preprint arXiv:2112.00114},
  year={2021}
}

@article{ouyang2022training,
  title={Training language models to follow instructions with human feedback},
  author={Ouyang, Long and Wu, Jeffrey and Jiang, Xu and Almeida, Diogo and Wainwright, Carroll and Mishkin, Pamela and Zhang, Chong and Agarwal, Sandhini and Slama, Katarina and Ray, Alex and others},
  journal={Advances in neural information processing systems},
  volume={35},
  pages={27730--27744},
  year={2022}
}

@article{zhao2023survey,
  title={A survey of large language models},
  author={Zhao, Wayne Xin and Zhou, Kun and Li, Junyi and Tang, Tianyi and Wang, Xiaolei and Hou, Yupeng and Min, Yingqian and Zhang, Beichen and Zhang, Junjie and Dong, Zican and others},
  journal={arXiv preprint arXiv:2303.18223},
  year={2023}
}

@article{lu2023emergent,
  title={Are Emergent Abilities in Large Language Models just In-Context Learning?},
  author={Lu, Sheng and Bigoulaeva, Irina and Sachdeva, Rachneet and Madabushi, Harish Tayyar and Gurevych, Iryna},
  journal={arXiv preprint arXiv:2309.01809},
  year={2023}
}

@article{liu2023emergent,
  title={Do emergent abilities exist in quantized large language models: An empirical study},
  author={Liu, Peiyu and Liu, Zikang and Gao, Ze-Feng and Gao, Dawei and Zhao, Wayne Xin and Li, Yaliang and Ding, Bolin and Wen, Ji-Rong},
  journal={arXiv preprint arXiv:2307.08072},
  year={2023}
}

@article{hopfield1982neural,
  title={Neural networks and physical systems with emergent collective computational abilities.},
  author={Hopfield, John J},
  journal={Proceedings of the national academy of sciences},
  volume={79},
  number={8},
  pages={2554--2558},
  year={1982},
  publisher={National Acad Sciences}
}

@article{du2024understanding,
  title={Understanding emergent abilities of language models from the loss perspective},
  author={Du, Zhengxiao and Zeng, Aohan and Dong, Yuxiao and Tang, Jie},
  journal={arXiv preprint arXiv:2403.15796},
  year={2024}
}

@article{schaeffer2024emergent,
  title={Are emergent abilities of large language models a mirage?},
  author={Schaeffer, Rylan and Miranda, Brando and Koyejo, Sanmi},
  journal={Advances in Neural Information Processing Systems},
  volume={36},
  year={2024}
}

@article{brown2020language,
  title={Language models are few-shot learners},
  author={Brown, Tom and Mann, Benjamin and Ryder, Nick and Subbiah, Melanie and Kaplan, Jared D and Dhariwal, Prafulla and Neelakantan, Arvind and Shyam, Pranav and Sastry, Girish and Askell, Amanda and others},
  journal={Advances in neural information processing systems},
  volume={33},
  pages={1877--1901},
  year={2020}
}

@article{touvron2023llama,
  title={Llama: Open and efficient foundation language models},
  author={Touvron, Hugo and Lavril, Thibaut and Izacard, Gautier and Martinet, Xavier and Lachaux, Marie-Anne and Lacroix, Timoth{\'e}e and Rozi{\`e}re, Baptiste and Goyal, Naman and Hambro, Eric and Azhar, Faisal and others},
  journal={arXiv preprint arXiv:2302.13971},
  year={2023}
}

@inproceedings{ganguli2022predictability,
  title={Predictability and surprise in large generative models},
  author={Ganguli, Deep and Hernandez, Danny and Lovitt, Liane and Askell, Amanda and Bai, Yuntao and Chen, Anna and Conerly, Tom and Dassarma, Nova and Drain, Dawn and Elhage, Nelson and others},
  booktitle={Proceedings of the 2022 ACM Conference on Fairness, Accountability, and Transparency},
  pages={1747--1764},
  year={2022}
}

@article{thoppilan2022lamda,
  title={Lamda: Language models for dialog applications},
  author={Thoppilan, Romal and De Freitas, Daniel and Hall, Jamie and Shazeer, Noam and Kulshreshtha, Apoorv and Cheng, Heng-Tze and Jin, Alicia and Bos, Taylor and Baker, Leslie and Du, Yu and others},
  journal={arXiv preprint arXiv:2201.08239},
  year={2022}
}

@article{huang2024c,
  title={C-eval: A multi-level multi-discipline chinese evaluation suite for foundation models},
  author={Huang, Yuzhen and Bai, Yuzhuo and Zhu, Zhihao and Zhang, Junlei and Zhang, Jinghan and Su, Tangjun and Liu, Junteng and Lv, Chuancheng and Zhang, Yikai and Fu, Yao and others},
  journal={Advances in Neural Information Processing Systems},
  volume={36},
  year={2024}
}

@inproceedings{DBLP:conf/iclr/HuSWALWWC22,
  author       = {Edward J. Hu and
                  Yelong Shen and
                  Phillip Wallis and
                  Zeyuan Allen{-}Zhu and
                  Yuanzhi Li and
                  Shean Wang and
                  Lu Wang and
                  Weizhu Chen},
  title        = {LoRA: Low-Rank Adaptation of Large Language Models},
  booktitle    = {The Tenth International Conference on Learning Representations, {ICLR}
                  2022, Virtual Event, April 25-29, 2022},
  publisher    = {OpenReview.net},
  year         = {2022},
  url          = {https://openreview.net/forum?id=nZeVKeeFYf9},
  bibsource    = {dblp computer science bibliography, https://dblp.org}
}

@article{wu2024inference,
  title={Inference scaling laws: An empirical analysis of compute-optimal inference for problem-solving with language models},
  author={Wu, Yangzhen and Sun, Zhiqing and Li, Shanda and Welleck, Sean and Yang, Yiming},
  journal={arXiv preprint arXiv:2408.00724},
  year={2024}
}

@misc{tanwar2023multilingualllmsbettercrosslingual,
      title={Multilingual LLMs are Better Cross-lingual In-context Learners with Alignment}, 
      author={Eshaan Tanwar and Subhabrata Dutta and Manish Borthakur and Tanmoy Chakraborty},
      year={2023},
      eprint={2305.05940},
      archivePrefix={arXiv},
      primaryClass={cs.CL},
      url={https://arxiv.org/abs/2305.05940}, 
}

@article{guo2024large,
  title={Large language model based multi-agents: A survey of progress and challenges},
  author={Guo, Taicheng and Chen, Xiuying and Wang, Yaqi and Chang, Ruidi and Pei, Shichao and Chawla, Nitesh V and Wiest, Olaf and Zhang, Xiangliang},
  journal={arXiv preprint arXiv:2402.01680},
  year={2024}
}

@article{chen2023agentverse,
  title={Agentverse: Facilitating multi-agent collaboration and exploring emergent behaviors in agents},
  author={Chen, Weize and Su, Yusheng and Zuo, Jingwei and Yang, Cheng and Yuan, Chenfei and Qian, Chen and Chan, Chi-Min and Qin, Yujia and Lu, Yaxi and Xie, Ruobing and others},
  journal={arXiv preprint arXiv:2308.10848},
  volume={2},
  number={4},
  pages={6},
  year={2023}
}

@article{hagendorff2024deception,
  title={Deception abilities emerged in large language models},
  author={Hagendorff, Thilo},
  journal={Proceedings of the National Academy of Sciences},
  volume={121},
  number={24},
  pages={e2317967121},
  year={2024},
  publisher={National Academy of Sciences}
}

@article{williams2024targeted,
  title={On targeted manipulation and deception when optimizing LLMs for user feedback},
  author={Williams, Marcus and Carroll, Micah and Narang, Adhyyan and Weisser, Constantin and Murphy, Brendan and Dragan, Anca},
  journal={arXiv preprint arXiv:2411.02306},
  year={2024}
}

@article{vercelli2024united,
  title={United Nations, artificial intelligences and regulations: analysis of the General Assembly AI Resolutions and the Final Report of the Advisory Body on AI},
  author={Vercelli, Ariel},
  year={2024}
}

@misc{blumenthal2024us,
  title={The US president’s executive order on artificial intelligence},
  author={Blumenthal, David},
  journal={NEJM AI},
  volume={1},
  number={2},
  pages={AIpc2300296},
  year={2024},
  publisher={Massachusetts Medical Society}
}

@article{eu2023aiact,
  title={Laying Down Harmonised Rules on Artificial Intelligence (Artificial Intelligence Act) and Amending Certain Union Legislative Acts},
  author={{European Union}},
  journal={Official Journal of the European Union},
  year={2023},
}

@misc{rae2022scaling,
      title={Scaling Language Models: Methods, Analysis \& Insights from Training Gopher}, 
      author={Jack W. Rae and Sebastian Borgeaud and Trevor Cai and Katie Millican and Jordan Hoffmann and Francis Song and John Aslanides and Sarah Henderson and Roman Ring and Susannah Young and Eliza Rutherford and Tom Hennigan and Jacob Menick and Albin Cassirer and Richard Powell and George van den Driessche and Lisa Anne Hendricks and Maribeth Rauh and Po-Sen Huang and Amelia Glaese and Johannes Welbl and Sumanth Dathathri and Saffron Huang and Jonathan Uesato and John Mellor and Irina Higgins and Antonia Creswell and Nat McAleese and Amy Wu and Erich Elsen and Siddhant Jayakumar and Elena Buchatskaya and David Budden and Esme Sutherland and Karen Simonyan and Michela Paganini and Laurent Sifre and Lena Martens and Xiang Lorraine Li and Adhiguna Kuncoro and Aida Nematzadeh and Elena Gribovskaya and Domenic Donato and Angeliki Lazaridou and Arthur Mensch and Jean-Baptiste Lespiau and Maria Tsimpoukelli and Nikolai Grigorev and Doug Fritz and Thibault Sottiaux and Mantas Pajarskas and Toby Pohlen and Zhitao Gong and Daniel Toyama and Cyprien de Masson d'Autume and Yujia Li and Tayfun Terzi and Vladimir Mikulik and Igor Babuschkin and Aidan Clark and Diego de Las Casas and Aurelia Guy and Chris Jones and James Bradbury and Matthew Johnson and Blake Hechtman and Laura Weidinger and Iason Gabriel and William Isaac and Ed Lockhart and Simon Osindero and Laura Rimell and Chris Dyer and Oriol Vinyals and Kareem Ayoub and Jeff Stanway and Lorrayne Bennett and Demis Hassabis and Koray Kavukcuoglu and Geoffrey Irving},
      year={2022},
      eprint={2112.11446},
      archivePrefix={arXiv},
      primaryClass={id='cs.CL' full_name='Computation and Language' is_active=True alt_name='cmp-lg' in_archive='cs' is_general=False description='Covers natural language processing. Roughly includes material in ACM Subject Class I.2.7. Note that work on artificial languages (programming languages, logics, formal systems) that does not explicitly address natural-language issues broadly construed (natural-language processing, computational linguistics, speech, text retrieval, etc.) is not appropriate for this area.'}
}

@article{bereska2024mechanistic,
  title={Mechanistic Interpretability for AI Safety--A Review},
  author={Bereska, Leonard and Gavves, Efstratios},
  journal={arXiv preprint arXiv:2404.14082},
  year={2024}
}

@article{snell2024scaling,
  title={Scaling llm test-time compute optimally can be more effective than scaling model parameters},
  author={Snell, Charlie and Lee, Jaehoon and Xu, Kelvin and Kumar, Aviral},
  journal={arXiv preprint arXiv:2408.03314},
  year={2024}
}

@article{shao2024deepseekmath,
  title={Deepseekmath: Pushing the limits of mathematical reasoning in open language models},
  author={Shao, Zhihong and Wang, Peiyi and Zhu, Qihao and Xu, Runxin and Song, Junxiao and Bi, Xiao and Zhang, Haowei and Zhang, Mingchuan and Li, YK and Wu, Y and others},
  journal={arXiv preprint arXiv:2402.03300},
  year={2024}
}

@misc{deepseekai2025deepseekr1incentivizingreasoningcapability,
      title={DeepSeek-R1: Incentivizing Reasoning Capability in LLMs via Reinforcement Learning}, 
      author={DeepSeek-AI and Daya Guo and Dejian Yang and Haowei Zhang and Junxiao Song and Ruoyu Zhang and Runxin Xu and Qihao Zhu and Shirong Ma and Peiyi Wang and Xiao Bi and Xiaokang Zhang and Xingkai Yu and Yu Wu and Z. F. Wu and Zhibin Gou and Zhihong Shao and Zhuoshu Li and Ziyi Gao and Aixin Liu and Bing Xue and Bingxuan Wang and Bochao Wu and Bei Feng and Chengda Lu and Chenggang Zhao and Chengqi Deng and Chenyu Zhang and Chong Ruan and Damai Dai and Deli Chen and Dongjie Ji and Erhang Li and Fangyun Lin and Fucong Dai and Fuli Luo and Guangbo Hao and Guanting Chen and Guowei Li and H. Zhang and Han Bao and Hanwei Xu and Haocheng Wang and Honghui Ding and Huajian Xin and Huazuo Gao and Hui Qu and Hui Li and Jianzhong Guo and Jiashi Li and Jiawei Wang and Jingchang Chen and Jingyang Yuan and Junjie Qiu and Junlong Li and J. L. Cai and Jiaqi Ni and Jian Liang and Jin Chen and Kai Dong and Kai Hu and Kaige Gao and Kang Guan and Kexin Huang and Kuai Yu and Lean Wang and Lecong Zhang and Liang Zhao and Litong Wang and Liyue Zhang and Lei Xu and Leyi Xia and Mingchuan Zhang and Minghua Zhang and Minghui Tang and Meng Li and Miaojun Wang and Mingming Li and Ning Tian and Panpan Huang and Peng Zhang and Qiancheng Wang and Qinyu Chen and Qiushi Du and Ruiqi Ge and Ruisong Zhang and Ruizhe Pan and Runji Wang and R. J. Chen and R. L. Jin and Ruyi Chen and Shanghao Lu and Shangyan Zhou and Shanhuang Chen and Shengfeng Ye and Shiyu Wang and Shuiping Yu and Shunfeng Zhou and Shuting Pan and S. S. Li and Shuang Zhou and Shaoqing Wu and Shengfeng Ye and Tao Yun and Tian Pei and Tianyu Sun and T. Wang and Wangding Zeng and Wanjia Zhao and Wen Liu and Wenfeng Liang and Wenjun Gao and Wenqin Yu and Wentao Zhang and W. L. Xiao and Wei An and Xiaodong Liu and Xiaohan Wang and Xiaokang Chen and Xiaotao Nie and Xin Cheng and Xin Liu and Xin Xie and Xingchao Liu and Xinyu Yang and Xinyuan Li and Xuecheng Su and Xuheng Lin and X. Q. Li and Xiangyue Jin and Xiaojin Shen and Xiaosha Chen and Xiaowen Sun and Xiaoxiang Wang and Xinnan Song and Xinyi Zhou and Xianzu Wang and Xinxia Shan and Y. K. Li and Y. Q. Wang and Y. X. Wei and Yang Zhang and Yanhong Xu and Yao Li and Yao Zhao and Yaofeng Sun and Yaohui Wang and Yi Yu and Yichao Zhang and Yifan Shi and Yiliang Xiong and Ying He and Yishi Piao and Yisong Wang and Yixuan Tan and Yiyang Ma and Yiyuan Liu and Yongqiang Guo and Yuan Ou and Yuduan Wang and Yue Gong and Yuheng Zou and Yujia He and Yunfan Xiong and Yuxiang Luo and Yuxiang You and Yuxuan Liu and Yuyang Zhou and Y. X. Zhu and Yanhong Xu and Yanping Huang and Yaohui Li and Yi Zheng and Yuchen Zhu and Yunxian Ma and Ying Tang and Yukun Zha and Yuting Yan and Z. Z. Ren and Zehui Ren and Zhangli Sha and Zhe Fu and Zhean Xu and Zhenda Xie and Zhengyan Zhang and Zhewen Hao and Zhicheng Ma and Zhigang Yan and Zhiyu Wu and Zihui Gu and Zijia Zhu and Zijun Liu and Zilin Li and Ziwei Xie and Ziyang Song and Zizheng Pan and Zhen Huang and Zhipeng Xu and Zhongyu Zhang and Zhen Zhang},
      year={2025},
      eprint={2501.12948},
      archivePrefix={arXiv},
      primaryClass={cs.CL},
      url={https://arxiv.org/abs/2501.12948}, 
}

@article{schaeffer2024has,
  title={Why Has Predicting Downstream Capabilities of Frontier AI Models with Scale Remained Elusive?},
  author={Schaeffer, Rylan and Schoelkopf, Hailey and Miranda, Brando and Mukobi, Gabriel and Madan, Varun and Ibrahim, Adam and Bradley, Herbie and Biderman, Stella and Koyejo, Sanmi},
  journal={arXiv preprint arXiv:2406.04391},
  year={2024}
}

@article{huang2024unified,
  title={Unified view of grokking, double descent and emergent abilities: A perspective from circuits competition},
  author={Huang, Yufei and Hu, Shengding and Han, Xu and Liu, Zhiyuan and Sun, Maosong},
  journal={arXiv preprint arXiv:2402.15175},
  year={2024}
}

@article{openai2023gpt,
  title={Gpt-4 technical report. arxiv 2303.08774},
  author={OpenAI, R},
  journal={View in Article},
  volume={2},
  number={5},
  year={2023}
}

@inproceedings{hu2023predicting,
  title={Predicting Emergent Abilities with Infinite Resolution Evaluation},
  author={Hu, Shengding and Liu, Xin and Han, Xu and Zhang, Xinrong and He, Chaoqun and Zhao, Weilin and Lin, Yankai and Ding, Ning and Ou, Zebin and Zeng, Guoyang and others},
  booktitle={The Twelfth International Conference on Learning Representations},
  year={2023}
}

@misc{chowdhery2022palm,
      title={PaLM: Scaling Language Modeling with Pathways}, 
      author={Aakanksha Chowdhery and Sharan Narang and Jacob Devlin and Maarten Bosma and Gaurav Mishra and Adam Roberts and Paul Barham and Hyung Won Chung and Charles Sutton and Sebastian Gehrmann and Parker Schuh and Kensen Shi and Sasha Tsvyashchenko and Joshua Maynez and Abhishek Rao and Parker Barnes and Yi Tay and Noam Shazeer and Vinodkumar Prabhakaran and Emily Reif and Nan Du and Ben Hutchinson and Reiner Pope and James Bradbury and Jacob Austin and Michael Isard and Guy Gur-Ari and Pengcheng Yin and Toju Duke and Anselm Levskaya and Sanjay Ghemawat and Sunipa Dev and Henryk Michalewski and Xavier Garcia and Vedant Misra and Kevin Robinson and Liam Fedus and Denny Zhou and Daphne Ippolito and David Luan and Hyeontaek Lim and Barret Zoph and Alexander Spiridonov and Ryan Sepassi and David Dohan and Shivani Agrawal and Mark Omernick and Andrew M. Dai and Thanumalayan Sankaranarayana Pillai and Marie Pellat and Aitor Lewkowycz and Erica Moreira and Rewon Child and Oleksandr Polozov and Katherine Lee and Zongwei Zhou and Xuezhi Wang and Brennan Saeta and Mark Diaz and Orhan Firat and Michele Catasta and Jason Wei and Kathy Meier-Hellstern and Douglas Eck and Jeff Dean and Slav Petrov and Noah Fiedel},
      year={2022},
      eprint={2204.02311},
      archivePrefix={arXiv},
      primaryClass={id='cs.CL' full_name='Computation and Language' is_active=True alt_name='cmp-lg' in_archive='cs' is_general=False description='Covers natural language processing. Roughly includes material in ACM Subject Class I.2.7. Note that work on artificial languages (programming languages, logics, formal systems) that does not explicitly address natural-language issues broadly construed (natural-language processing, computational linguistics, speech, text retrieval, etc.) is not appropriate for this area.'}
}

@misc{johnson2006emergent,
  title={What are emergent properties and how do they affect the engineering of complex systems?},
  author={Johnson, Christopher W},
  journal={Reliability Engineering \& System Safety},
  volume={91},
  number={12},
  pages={1475--1481},
  year={2006},
  publisher={Elsevier}
}

@article{hoffmann2022training,
  title={Training compute-optimal large language models},
  author={Hoffmann, Jordan and Borgeaud, Sebastian and Mensch, Arthur and Buchatskaya, Elena and Cai, Trevor and Rutherford, Eliza and Casas, Diego de Las and Hendricks, Lisa Anne and Welbl, Johannes and Clark, Aidan and others},
  journal={arXiv preprint arXiv:2203.15556} ,
  year={2022}
}

@misc{steinhardt2022emergent,
  title        = {Future ML Systems Will Be Qualitatively Different},
  author       = {Jacob Steinhardt},
  year         = 2022,
  howpublished = {\url{https://bounded-regret.ghost.io/future-ml-systems-will-be-qualitatively-different/}}
}

@article{anderson1972more,
  title={More Is Different: Broken symmetry and the nature of the hierarchical structure of science.},
  author={Anderson, Philip W},
  journal={Science},
  volume={177},
  number={4047},
  pages={393--396},
  year={1972},
  publisher={American Association for the Advancement of Science}
}

@article{shin2022effect,
  title={On the effect of pretraining corpora on in-context learning by a large-scale language model},
  author={Shin, Seongjin and Lee, Sang-Woo and Ahn, Hwijeen and Kim, Sungdong and Kim, HyoungSeok and Kim, Boseop and Cho, Kyunghyun and Lee, Gichang and Park, Woomyoung and Ha, Jung-Woo and others},
  journal={arXiv preprint arXiv:2204.13509},
  year={2022}
}

@article{yadlowsky2023pretraining,
  title={Pretraining data mixtures enable narrow model selection capabilities in transformer models},
  author={Yadlowsky, Steve and Doshi, Lyric and Tripuraneni, Nilesh},
  journal={arXiv preprint arXiv:2311.00871},
  year={2023}
}

@article{wies2024learnability,
  title={The learnability of in-context learning},
  author={Wies, Noam and Levine, Yoav and Shashua, Amnon},
  journal={Advances in Neural Information Processing Systems},
  volume={36},
  year={2024}
}

@article{raventos2024pretraining,
  title={Pretraining task diversity and the emergence of non-bayesian in-context learning for regression},
  author={Ravent{\'o}s, Allan and Paul, Mansheej and Chen, Feng and Ganguli, Surya},
  journal={Advances in Neural Information Processing Systems},
  volume={36},
  year={2024}
}

@article{ding2023causallm,
  title={CausalLM is not optimal for in-context learning},
  author={Ding, Nan and Levinboim, Tomer and Wu, Jialin and Goodman, Sebastian and Soricut, Radu},
  journal={arXiv preprint arXiv:2308.06912},
  year={2023}
}

@inproceedings{zhao2021calibrate,
  title={Calibrate before use: Improving few-shot performance of language models},
  author={Zhao, Zihao and Wallace, Eric and Feng, Shi and Klein, Dan and Singh, Sameer},
  booktitle={International conference on machine learning},
  pages={12697--12706},
  year={2021},
  organization={PMLR}
}

@article{lu2021fantastically,
  title={Fantastically ordered prompts and where to find them: Overcoming few-shot prompt order sensitivity},
  author={Lu, Yao and Bartolo, Max and Moore, Alastair and Riedel, Sebastian and Stenetorp, Pontus},
  journal={arXiv preprint arXiv:2104.08786},
  year={2021}
}

@article{liu2021makes,
  title={What Makes Good In-Context Examples for GPT-$3 $?},
  author={Liu, Jiachang and Shen, Dinghan and Zhang, Yizhe and Dolan, Bill and Carin, Lawrence and Chen, Weizhu},
  journal={arXiv preprint arXiv:2101.06804},
  year={2021}
}

@article{wu2022self,
  title={Self-adaptive in-context learning: An information compression perspective for in-context example selection and ordering},
  author={Wu, Zhiyong and Wang, Yaoxiang and Ye, Jiacheng and Kong, Lingpeng},
  journal={arXiv preprint arXiv:2212.10375},
  year={2022}
}

@article{sorensen2022information,
  title={An information-theoretic approach to prompt engineering without ground truth labels},
  author={Sorensen, Taylor and Robinson, Joshua and Rytting, Christopher Michael and Shaw, Alexander Glenn and Rogers, Kyle Jeffrey and Delorey, Alexia Pauline and Khalil, Mahmoud and Fulda, Nancy and Wingate, David},
  journal={arXiv preprint arXiv:2203.11364},
  year={2022}
}

@article{gonen2022demystifying,
  title={Demystifying prompts in language models via perplexity estimation},
  author={Gonen, Hila and Iyer, Srini and Blevins, Terra and Smith, Noah A and Zettlemoyer, Luke},
  journal={arXiv preprint arXiv:2212.04037},
  year={2022}
}

@article{li2023finding,
  title={Finding supporting examples for in-context learning},
  author={Li, Xiaonan and Qiu, Xipeng},
  journal={CoRR},
  year={2023}
}

@article{rubin2021learning,
  title={Learning to retrieve prompts for in-context learning},
  author={Rubin, Ohad and Herzig, Jonathan and Berant, Jonathan},
  journal={arXiv preprint arXiv:2112.08633},
  year={2021}
}

@article{li2023unified,
  title={Unified demonstration retriever for in-context learning},
  author={Li, Xiaonan and Lv, Kai and Yan, Hang and Lin, Tianyang and Zhu, Wei and Ni, Yuan and Xie, Guotong and Wang, Xiaoling and Qiu, Xipeng},
  journal={arXiv preprint arXiv:2305.04320},
  year={2023}
}

@inproceedings{ye2023compositional,
  title={Compositional exemplars for in-context learning},
  author={Ye, Jiacheng and Wu, Zhiyong and Feng, Jiangtao and Yu, Tao and Kong, Lingpeng},
  booktitle={International Conference on Machine Learning},
  pages={39818--39833},
  year={2023},
  organization={PMLR}
}

@article{zhang2022active,
  title={Active example selection for in-context learning},
  author={Zhang, Yiming and Feng, Shi and Tan, Chenhao},
  journal={arXiv preprint arXiv:2211.04486},
  year={2022}
}

@misc{gottweis2025towards,
    title = {Towards an AI co-scientist},
    year={2025},
author = {Gottweis, Juraj and Weng, Wei-Hung and Daryin, Alexander and Tu, Tao  and Palepu, Anil and Sirkovic, Petar and Myaskovsky, Artiom and Weissenberger, Felix and Rong, Keran and Tanno, Ryutaro and Saab, Khaled and Popovici, Dan and Blum, Jacob and Zhang, Fan and Chou, Katherine and Hassidim, Avinatan and Gokturk, Burak and Vahdat, Amin and Kohli, Pushmeet and Matias, Yossi and Carroll, Andrew and Kulkarni, Kavita and Tomasev, Nenad and Dhillon, Vikram and Dhaval Vaishnav, Eeshit and Lee, Byron and R.D. Costa, Tiago and Penadés, José R. and Peltz, Gary and Xu, Yunhan and Pawlosky, Annalisa and  Karthikesalingam, Alan and Natarajan, Vivek},
    url = {https://storage.googleapis.com/coscientist_paper/ai_coscientist.pdf}
}

@book{bostrom2024superintelligence,
  title={Superintelligence},
  author={Bostrom, Nick},
  year={2024},
  publisher={Dunod}
}

@article{kim2022self,
  title={Self-generated in-context learning: Leveraging auto-regressive language models as a demonstration generator},
  author={Kim, Hyuhng Joon and Cho, Hyunsoo and Kim, Junyeob and Kim, Taeuk and Yoo, Kang Min and Lee, Sang-goo},
  journal={arXiv preprint arXiv:2206.08082},
  year={2022}
}

@article{yang2023auto,
  title={Auto-ICL: In-Context Learning without Human Supervision},
  author={Yang, Jinghan and Ma, Shuming and Wei, Furu},
  journal={arXiv preprint arXiv:2311.09263},
  year={2023}
}

@article{hao2022structured,
  title={Structured prompting: Scaling in-context learning to 1,000 examples},
  author={Hao, Yaru and Sun, Yutao and Dong, Li and Han, Zhixiong and Gu, Yuxian and Wei, Furu},
  journal={arXiv preprint arXiv:2212.06713},
  year={2022}
}

@article{liu2024let,
  title={Let's Learn Step by Step: Enhancing In-Context Learning Ability with Curriculum Learning},
  author={Liu, Yinpeng and Liu, Jiawei and Shi, Xiang and Cheng, Qikai and Huang, Yong and Lu, Wei},
  journal={arXiv preprint arXiv:2402.10738},
  year={2024}
}

@article{honovich2022instruction,
  title={Instruction induction: From few examples to natural language task descriptions},
  author={Honovich, Or and Shaham, Uri and Bowman, Samuel R and Levy, Omer},
  journal={arXiv preprint arXiv:2205.10782},
  year={2022}
}

@article{zhou2022large,
  title={Large language models are human-level prompt engineers},
  author={Zhou, Yongchao and Muresanu, Andrei Ioan and Han, Ziwen and Paster, Keiran and Pitis, Silviu and Chan, Harris and Ba, Jimmy},
  journal={arXiv preprint arXiv:2211.01910},
  year={2022}
}

@article{wang2022self,
  title={Self-instruct: Aligning language models with self-generated instructions},
  author={Wang, Yizhong and Kordi, Yeganeh and Mishra, Swaroop and Liu, Alisa and Smith, Noah A and Khashabi, Daniel and Hajishirzi, Hannaneh},
  journal={arXiv preprint arXiv:2212.10560},
  year={2022}
}

@article{olsson2022context,
  title={In-context learning and induction heads},
  author={Olsson, Catherine and Elhage, Nelson and Nanda, Neel and Joseph, Nicholas and DasSarma, Nova and Henighan, Tom and Mann, Ben and Askell, Amanda and Bai, Yuntao and Chen, Anna and others},
  journal={arXiv preprint arXiv:2209.11895},
  year={2022}
}

@article{wang2023label,
  title={Label words are anchors: An information flow perspective for understanding in-context learning},
  author={Wang, Lean and Li, Lei and Dai, Damai and Chen, Deli and Zhou, Hao and Meng, Fandong and Zhou, Jie and Sun, Xu},
  journal={arXiv preprint arXiv:2305.14160},
  year={2023}
}

@article{bietti2024birth,
  title={Birth of a transformer: A memory viewpoint},
  author={Bietti, Alberto and Cabannes, Vivien and Bouchacourt, Diane and Jegou, Herve and Bottou, Leon},
  journal={Advances in Neural Information Processing Systems},
  volume={36},
  year={2024}
}

@inproceedings{irie2022dual,
  title={The dual form of neural networks revisited: Connecting test time predictions to training patterns via spotlights of attention},
  author={Irie, Kazuki and Csord{\'a}s, R{\'o}bert and Schmidhuber, J{\"u}rgen},
  booktitle={International Conference on Machine Learning},
  pages={9639--9659},
  year={2022},
  organization={PMLR}
}

@article{elhage2021mathematical,
  title={A mathematical framework for transformer circuits},
  author={Elhage, Nelson and Nanda, Neel and Olsson, Catherine and Henighan, Tom and Joseph, Nicholas and Mann, Ben and Askell, Amanda and Bai, Yuntao and Chen, Anna and Conerly, Tom and others},
  journal={Transformer Circuits Thread},
  volume={1},
  number={1},
  pages={12},
  year={2021}
}

@article{todd2023function,
  title={Function vectors in large language models},
  author={Todd, Eric and Li, Millicent L and Sharma, Arnab Sen and Mueller, Aaron and Wallace, Byron C and Bau, David},
  journal={arXiv preprint arXiv:2310.15213},
  year={2023}
}

@article{xie2021explanation,
  title={An explanation of in-context learning as implicit bayesian inference},
  author={Xie, Sang Michael and Raghunathan, Aditi and Liang, Percy and Ma, Tengyu},
  journal={arXiv preprint arXiv:2111.02080},
  year={2021}
}

@article{min2022rethinking,
  title={Rethinking the role of demonstrations: What makes in-context learning work?},
  author={Min, Sewon and Lyu, Xinxi and Holtzman, Ari and Artetxe, Mikel and Lewis, Mike and Hajishirzi, Hannaneh and Zettlemoyer, Luke},
  journal={arXiv preprint arXiv:2202.12837},
  year={2022}
}

@article{chan2022data,
  title={Data distributional properties drive emergent in-context learning in transformers},
  author={Chan, Stephanie and Santoro, Adam and Lampinen, Andrew and Wang, Jane and Singh, Aaditya and Richemond, Pierre and McClelland, James and Hill, Felix},
  journal={Advances in Neural Information Processing Systems},
  volume={35},
  pages={18878--18891},
  year={2022}
}

@article{sanh2021multitask,
  title={Multitask prompted training enables zero-shot task generalization},
  author={Sanh, Victor and Webson, Albert and Raffel, Colin and Bach, Stephen H and Sutawika, Lintang and Alyafeai, Zaid and Chaffin, Antoine and Stiegler, Arnaud and Scao, Teven Le and Raja, Arun and others},
  journal={arXiv preprint arXiv:2110.08207},
  year={2021}
}

@article{razeghi2022impact,
  title={Impact of pretraining term frequencies on few-shot reasoning},
  author={Razeghi, Yasaman and Logan IV, Robert L and Gardner, Matt and Singh, Sameer},
  journal={arXiv preprint arXiv:2202.07206},
  year={2022}
}

@article{hahn2023theory,
  title={A theory of emergent in-context learning as implicit structure induction},
  author={Hahn, Michael and Goyal, Navin},
  journal={arXiv preprint arXiv:2303.07971},
  year={2023}
}

@article{arora2023theory,
  title={A theory for emergence of complex skills in language models},
  author={Arora, Sanjeev and Goyal, Anirudh},
  journal={arXiv preprint arXiv:2307.15936},
  year={2023}
}

@article{devlin2018bert,
  title={Bert: Pre-training of deep bidirectional transformers for language understanding},
  author={Devlin, Jacob and Chang, Ming-Wei and Lee, Kenton and Toutanova, Kristina},
  journal={arXiv preprint arXiv:1810.04805},
  year={2018}
}

@article{kaplan2020scaling,
  title={Scaling laws for neural language models},
  author={Kaplan, Jared and McCandlish, Sam and Henighan, Tom and Brown, Tom B and Chess, Benjamin and Child, Rewon and Gray, Scott and Radford, Alec and Wu, Jeffrey and Amodei, Dario},
  journal={arXiv preprint arXiv:2001.08361},
  year={2020}
}

@article{li2024memory,
  title={Memory, Consciousness and Large Language Model},
  author={Li, Jitang and Li, Jinzheng},
  journal={arXiv preprint arXiv:2401.02509},
  year={2024}
}

@article{DBLP:journals/corr/abs-2303-07103,
  author       = {David J. Chalmers},
  title        = {Could a Large Language Model be Conscious?},
  journal      = {CoRR},
  volume       = {abs/2303.07103},
  year         = {2023},
  url          = {https://doi.org/10.48550/arXiv.2303.07103},
  doi          = {10.48550/ARXIV.2303.07103},
  eprinttype    = {arXiv},
  eprint       = {2303.07103},
  bibsource    = {dblp computer science bibliography, https://dblp.org}
}

@article{srivastava2022beyond,
  title={Beyond the imitation game: Quantifying and extrapolating the capabilities of language models},
  author={Srivastava, Aarohi and Rastogi, Abhinav and Rao, Abhishek and Shoeybi, Mohammad and Casper, Bryan and Patwary, Mostofa and Le, Quoc V and Bittorf, Victor and Chung, Jun and Pham, Phu and others},
  journal={arXiv preprint arXiv:2206.04615} ,
  year={2022}
}


\end{document}